\PassOptionsToPackage{dvipsnames}{xcolor} %
\documentclass[conference, 9pt]{IEEEtran}
\pdfoutput=1
\IEEEoverridecommandlockouts
\usepackage[style=ieee, citestyle=numeric, sortcites=true, backend=bibtex, maxbibnames=99]{biblatex}
\usepackage[T1]{fontenc} %
\usepackage{fancyhdr}

\usepackage{lmodern}
\usepackage{amsmath,amssymb,amsfonts}
\usepackage{algorithm,algorithmicx,algpseudocode}
\usepackage{listings}
\usepackage{graphicx}
\usepackage{booktabs}
\usepackage{textcomp}
\usepackage{xcolor}
\usepackage{tikz}
\usepackage{comment}
\usepackage{tabularx, supertabular}
\usepackage{hyperref}
\usepackage{subcaption}
\usepackage{longtable}
\usepackage{sidecap}    \sidecaptionvpos{figure}{t}
\usepackage{url, hyperref}                  %
\hypersetup{
    colorlinks, linkcolor=RoyalBlue, citecolor=RoyalBlue, urlcolor=RoyalBlue,
    allcolors=RoyalBlue,
    pdfborder={0 0 0}}
\usepackage{xcolor}

\usepackage{pifont}%

\newcommand{\TwoLineTabTitle}[3][c]{\begin{tabular}{@{}#1@{}} #2 \\[-.5ex] #3\end{tabular}}
\newcommand{\ThreeLineTabTitle}[4][c]{\begin{tabular}{@{}#1@{}} #2 \\[-.5ex] #3\\[-.5ex] #4\end{tabular}}
\newcommand{\TabHighlightData}[1]{\color{black}\bfseries #1}
\newcommand{\NotTabHighlightData}[1]{\color{gray}\bfseries #1}

\def\BibTeX{{\rm B\kern-.05em{\sc i\kern-.025em b}\kern-.08em
    T\kern-.1667em\lower.7ex\hbox{E}\kern-.125emX}}

\begin{document}

\title{Accelerating Communication in Deep Learning Recommendation Model Training with Dual-Level Adaptive Lossy Compression}

\author{\normalsize\null\\[-4.2ex] %
Hao Feng\IEEEauthorrefmark{1},
Boyuan Zhang\IEEEauthorrefmark{1}\thanks{Hao Feng and Boyuan Zhang contribute equally in this work.},
Fanjiang Ye\IEEEauthorrefmark{1},
Min Si\IEEEauthorrefmark{2},
Ching-Hsiang Chu\IEEEauthorrefmark{2},
Jiannan Tian\IEEEauthorrefmark{1}\thanks{Jiannan Tian's participation in this work ended on 03/26/2024.},
Chunxing Yin\IEEEauthorrefmark{2},\\
Summer Deng\IEEEauthorrefmark{2},
Yuchen Hao\IEEEauthorrefmark{2},
Pavan Balaji\IEEEauthorrefmark{2},
Tong Geng\IEEEauthorrefmark{3},
Dingwen Tao\IEEEauthorrefmark{4}\thanks{Dingwen Tao is the corresponding author.}\\[.9ex]  %
\IEEEauthorrefmark{1} Indiana University, Bloomington, IN, USA;\ \ \texttt{\string{\,haofeng, bozhan, fanjye, jti1\,\string}@iu.edu}\\
\IEEEauthorrefmark{2}
Meta, Menlo Park, CA, USA;\ \ \texttt{\string{\,msi, chchu, cyin9, summerdeng, haoyc, pavanbalaji\,\string}@meta.com} \\
\IEEEauthorrefmark{3}
University of Rochester, Rochester, NY, USA;\ \ \texttt{tgeng@ur.rochester.edu}\\
\IEEEauthorrefmark{4}
SKLP, Institute of Computing Technology, Chinese Academy of Sciences, China;\ \ \texttt{taodingwen@ict.ac.cn}\\[-3ex]  %
}

\maketitle %
\thispagestyle{fancy}
\lhead{}
\rhead{}
\chead{}
\lfoot{\footnotesize SC24, November 17--22, 2024, Atlanta, GA, USA \newline 979-8-3503-5291-7/24/\$31.00 \copyright 2024 IEEE} 
\rfoot{}
\cfoot{}
\renewcommand{\headrulewidth}{0pt} \renewcommand{\footrulewidth}{0pt}

\begin{abstract}

DLRM is a state-of-the-art recommendation system model that has gained widespread adoption across various industry applications. The large size of DLRM models, however, necessitates the use of multiple devices/GPUs for efficient training. A significant bottleneck in this process is the time-consuming all-to-all communication required to collect embedding data from all devices. To mitigate this, we introduce a method that employs error-bounded lossy compression to reduce the communication data size and accelerate DLRM training. We develop a novel error-bounded lossy compression algorithm, informed by an in-depth analysis of embedding data features, to achieve high compression ratios. Moreover, we introduce a dual-level adaptive strategy for error-bound adjustment, spanning both table-wise and iteration-wise aspects, to balance the compression benefits with the potential impacts on accuracy. We further optimize our compressor for PyTorch tensors on GPUs, minimizing compression overhead. Evaluation shows that our method achieves a 1.38$\times$ training speedup with a minimal accuracy impact.

\end{abstract}

\section{Introduction}
Deep Learning Recommendation Models (DLRMs) have significantly risen to prominence in both research and industry sectors in recent years. These models integrate sparse input embedding learning with neural network architectures, marking a notable advance over traditional collaborative filtering-based recommendation systems \cite{dlrm}. DLRMs have been successfully implemented in various industry applications, including product recommendations system by Amazon \cite{ma2020temporal}, personalized advertising by Google \cite{cheng2016wide}, and e-commerce service by Alibaba \cite{wang2018billion}. As a result, they constitute a significant portion of deep learning applications across multiple industries.

DLRMs are uniquely designed to process high-dimensional categorical features, typically represented by one- or multi-hot vectors matching the size of the category, which leads to significant data sparsity. To efficiently manage this, DLRMs utilize embedding tables that transform these high-dimensional sparse vectors into lower-dimensional, dense vector representations. In a typical DLRM architecture, dense features are processed through a multi-layer perceptron (MLP), combined with sparse embedding lookups in a feature interaction module, and then fed into the top MLP. This process culminates in generating a click-through rate (CTR) prediction. Such a structure elegantly combines sparse and dense data processing, underscoring the complexity and challenges associated with the efficient implementation and scaling of DLRMs.

A critical challenge in deploying large-scale DLRMs lies in managing the massive size of embedding tables, which can extend to terabytes, far exceeding the memory capacity of a single GPU. To address this issue, hybrid-parallel distributed training systems are widely employed. In these systems, MLP layers are replicated across multiple GPUs for data-parallel training, while embedding tables are partitioned and distributed for model-parallel training. This setup necessitates the use of collective communication primitives for synchronization across all GPUs. Specifically, the partitioning of sparse embedding tables requires nodes to aggregate sparse embedding lookups during forward passes and their corresponding gradients during backward passes. Consequently, \textbf{all-to-all} communication is utilized in both forward and backward passes for synchronizing sparse lookups and gradients, while all-reduce is employed for synchronizing dense/MLP gradients during the backward pass.

\begin{figure}[t] 
\centering
\includegraphics[width=\linewidth]{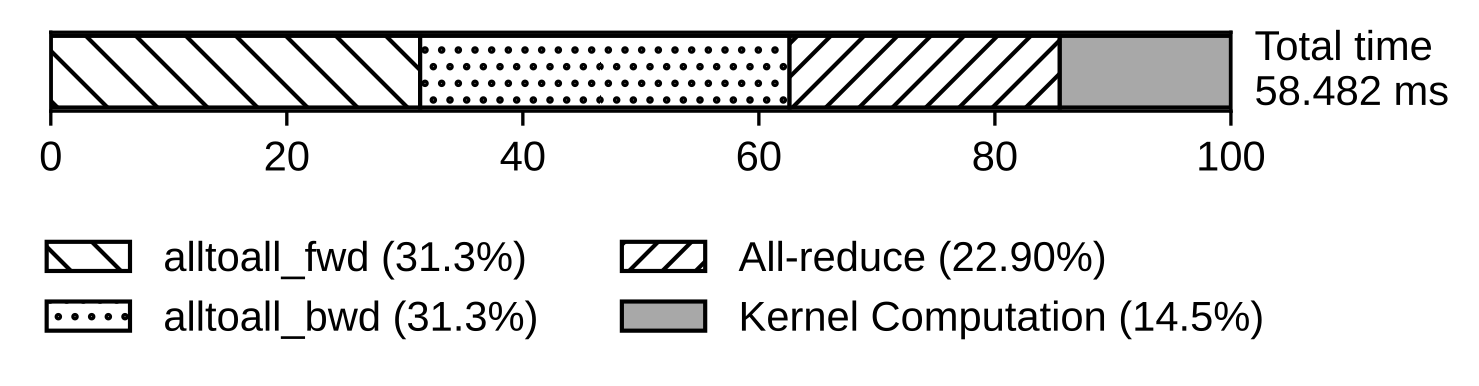} 
\caption{Performance profiling of DLRM training with 32 GPUs.}
\label{32GPU_profiling}
\end{figure}

Communication for synchronizing embedding lookups and gradients across all GPUs during each mini-batch iteration significantly contributes to the overall training time. For example, Figure \ref{32GPU_profiling} shows that all-to-all communication accounts for more than 60\% of the total training time for DLRM on an 8-node, 32 A100 GPUs cluster (connected through a Slingshot 10 interconnect \cite{DBLP:conf/sc/SensiGMRH20}). Consequently, various studies have been conducted to address these communication challenges. 

One method involves the application of low-bit quantization (e.g., FP16, FP8) to represent embedding tables \cite{DBLP:journals/corr/abs-2010-11305}. However, quantization has two primary limitations: \ding{182} Its capacity for data reduction (e.g., 2$\times$) is relatively limited. \ding{183} While quantization is viable for inference, training with a quantized embedding table often results in significant accuracy losses \cite{4bitinfer}. Another approach is the use of lossless compression to compress embedding lookups just before the all-to-all communication \cite{DBLP:conf/mlhpc-ws/PummaV21}. However, this method also faces challenges due to the sparse and random nature of embedding lookups and the mantissa of floating-point data, which limits the achievable compression ratio.

Unlike quantization and lossless compression approaches, error-bounded lossy compression achieves a significantly higher data reduction ratio while maintaining strict error control in the reconstructed data. However, effectively employing lossy compression in DLRM training necessitates addressing several key challenges: \ding{182} \textbf{Low Compression Ratio:} Existing error-bounded lossy compression methods, such as SZ \cite{sz16} and ZFP \cite{zfp}, also face the challenge of achieving a high compression ratio on embedding lookups (which will be explained in Section \ref{sec:design}). Thus, it is essential to develop a lossy compression algorithm optimized for embedding lookups.
\ding{183} \textbf{High Compression Overhead:} Compression for every all-to-all communication at each iteration introduces a high compression overhead; thus, implementing an efficient compression algorithm on GPUs and seamlessly integrating it into both the communication and DLRM computation workflows is vital, ensuring minimal performance overhead.
\ding{184} \textbf{Error Propagation:} Lossy compression introduces errors to the reconstructed embedding lookups after all-to-all communication. Thus, developing a strategy for adaptively controlling error bounds across different embedding tables and training iterations is critical to ensure an acceptable impact on accuracy.

To address these challenges, we introduce a highly efficient approach to accelerate communication in DLRM training through the use of error-bounded lossy compression, deeply optimizing and adaptively applying it. 

Our key contributions include:
\begin{itemize}
\item We introduce a novel hybrid compression method for embedding lookups/vectors, consisting of two algorithms: a newly developed LZ compression algorithm for embedding vectors and an optimized entropy-based Huffman compression algorithm for vector elements.
\item We develop a two-level adaptive strategy for error-bound adjustment for different embedding tables and training iterations, aiming to maintain relatively large error bounds (for higher compression ratios) while minimizing the impact on accuracy.
\item We optimize our compression method on modern GPUs, enabling parallel compression of multiple tensors into a single compressed tensor, effectively minimizing data movements and kernel launches.
\item We evaluate our method using three widely used DLRM datasets with up to 32 GPUs and demonstrate that our method significantly accelerates all-to-all communication in DLRM training by 8.6$\times$, with an accuracy loss of less than 0.02\%—well within the tolerable level.
\end{itemize}

\section{Background and Problem Statement}
\label{sec:background}

\subsection{Deep Learning Recommendation Model (DLRM)}

DLRM is a widely used recommendation model, which is designed to utilize both categorical and numerical inputs for personalized recommendations. We will discuss the architecture, pipeline, and large-scale training of DLRM.

\begin{figure}[t] 
\centering 
\includegraphics[width=\linewidth]{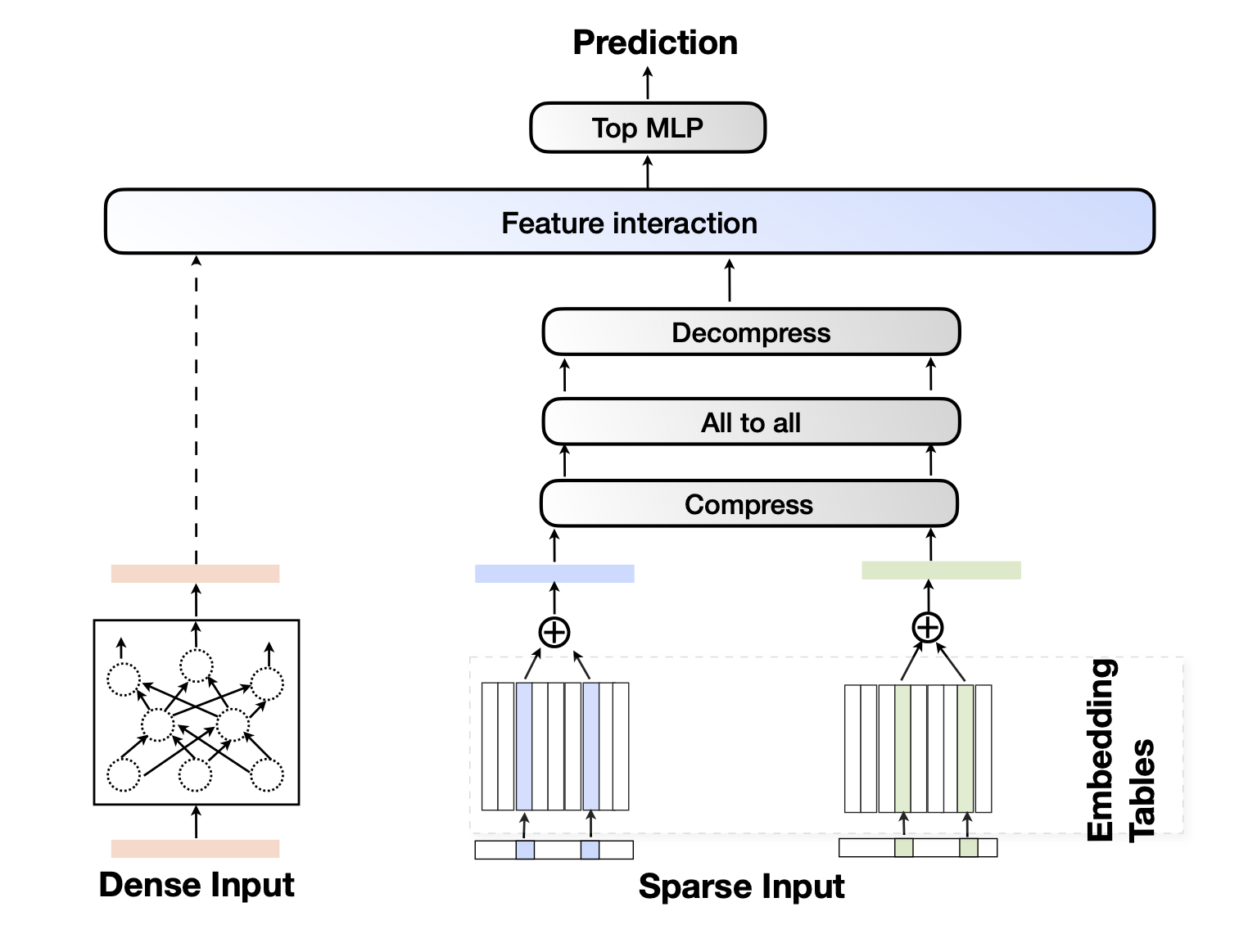}
\caption{Illustration of DLRM architecture.} 
\label{fig:dlrm-arch}
\end{figure}

\textbf{DLRM Architecture}
Generally, DLRM comprises three components: Embedding tables, Interaction Module, and MLP. The architecture is shown in Figure \ref{fig:dlrm-arch}. Embedding tables in DLRM process categorical data by looking up each categorical feature and mapping it into an embedding vector representation as the output vector. The Interaction Module will apply input vectors from the embedding tables and the Bottom MLP, and perform interactions on them to generate a new output vector. There are two MLPs in DLRM: the Bottom MLP and the Top MLP. The Bottom MLP transforms dense features to match the length of embedding vectors. The Top MLP applies the concatenated data as input and calculates the Click-Through Rate (CTR) as output.

\textbf{DLRM Training Parallelisms and Bottlenecks}
Training DLRM involves both model and data parallelism to manage its diverse computational needs efficiently. Model parallelism is crucial for handling the large Embedding Tables (EMBs) distributed across different devices due to their size, while data parallelism is applied to the Multilayer Perceptrons (MLPs), which, despite requiring full access every epoch, consume a relatively small amount of memory. This dual approach allows for the division of a global batch of embedding vectors into smaller local batches, facilitating interaction operations and subsequent processing by the Top MLP through an all-to-all communication pattern. This setup is particularly effective for implementing various second-order interaction methods, including dot products between pairs of embedding vectors and dense features, which are then concatenated with dense features for input to the Top MLP, ultimately enabling classification. During backward propagation, a symmetrical all-to-all communication redistributes gradients back to their respective devices for updating the Embedding tables and Bottom MLP, reflecting the forward phase's operations. 

The bottlenecks across DLRM's components vary: embedding layers are bandwidth-dominated due to high-bandwidth memory access requirements for lookup operations, MLP layers are computation-dominated, and the feature interaction layer is communication-dominated. Our profiling indicates that DLRM training is notably communication-intensive, underscoring the necessity of optimizing these parallelism strategies for large-scale training efficiency (see details in Section \ref{sec:evaluation}).

\subsection{Floating-point Data Compression}

There are two principal classes of data compression: lossless and lossy. While lossless compression preserves data integrity perfectly, lossy compression achieves significantly higher compression ratios at the cost of acceptable accuracy loss. Traditional lossy compressors, such as JPEG \cite{jpeg} for images and MPEG \cite{mpeg} for videos, are designed with human perception in mind, lacking precise error-control mechanisms for scientific post-analysis.

A new generation of lossy compression techniques for scientific data, particularly floating-point data, has emerged, exemplified by SZ \cite{sz16, sz17, sz3liang}, ZFP \cite{zfp}, 
and TTHRESH \cite{BLP:19}.
Unlike their counterparts for media, these scientific data compressors offer strict error-controlling schemes, enabling users to manage accuracy loss both in the reconstructed data and during post-analysis.

With the proliferation of GPU-based HPC systems and applications, compressors like SZ and ZFP
have introduced GPU-optimized versions (i.e., cuSZ \cite{cusz2020} and cuZFP \cite{zfp}),
delivering significantly enhanced throughput compared to their CPU-based implementations. ZFP, a transform-based compressor, allows users to set a desired bitrate, while SZ, a prediction-based compressor,
enables the specification of a maximum tolerable error. ZFP in fixed-rate mode tends to offer consistently higher throughput, whereas SZ 
in error-bounded mode achieves superior compression ratios. 

\subsection{Problem Statement}

\textbf{Dominance of Communication Overhead.} As illustrated in Figure \ref{32GPU_profiling}, all-to-all communication accounts for over 60\% of DLRM's total training time, establishing communication as the bottleneck rather than computation. This bottleneck is exacerbated when training DLRM with datasets of various sizes. For instance, the Criteo Kaggle dataset, with sparse feature lengths of 32, generates more than 121 GB of lookup data per epoch. This figure escalates dramatically with larger datasets, such as the Criteo Terabyte dataset, which can accumulate up to terabytes of lookup data per epoch. The scale increases further with industry-level recommendation models, often exceeding multiple terabytes, necessitating larger volumes of training data and distributed systems of larger scales for parameter storage. This significantly increases communication data across devices, highlighting the urgent need to reduce communication data volume in DLRM training.

However, as aforementioned, there are several challenges to addressing this issue due to the limitations of directly applying current error-bounded lossy compression techniques. To effectively tackle these hurdles, this paper focuses on the following key research questions:
\ding{182} \textbf{Error Bound Configuration:} It's essential to determine the optimal error bounds for various embedding tables across different training iterations to maintain training accuracy while enhancing compression ratios.
\ding{183} \textbf{DLRM-specific Lossy Compression Algorithm:} There's a critical need to devise a compression algorithm uniquely suited for DLRM's embedding tables, aiming for elevated compression ratios without substantially compromising data integrity.
\ding{184} \textbf{GPU Compression Performance Optimization:} Optimizing the GPU execution of our specialized compression and ensuring its smooth incorporation within DLRM training is crucial for improving overall training performance.

\section{System Design}
\label{sec:design}
In this section, we discuss our approach to accelerating DLRM training, divided into four main parts. First, we provide an overview of our proposed training pipeline that incorporates lossy compression in Section \ref{sub1:pipeline}. Second, we share our observations of DLRM data features 
in Section \ref{sub2:observation}. Following this, we explain how we dynamically adjust our error bound during training. Lastly, we introduce our optimized compression algorithm designed to enhance performance.

\subsection{Overview of DLRM Training Pipeline with Compression}
\label{sub1:pipeline} 
\begin{figure}[t]%
\centering
\includegraphics[width=\linewidth]{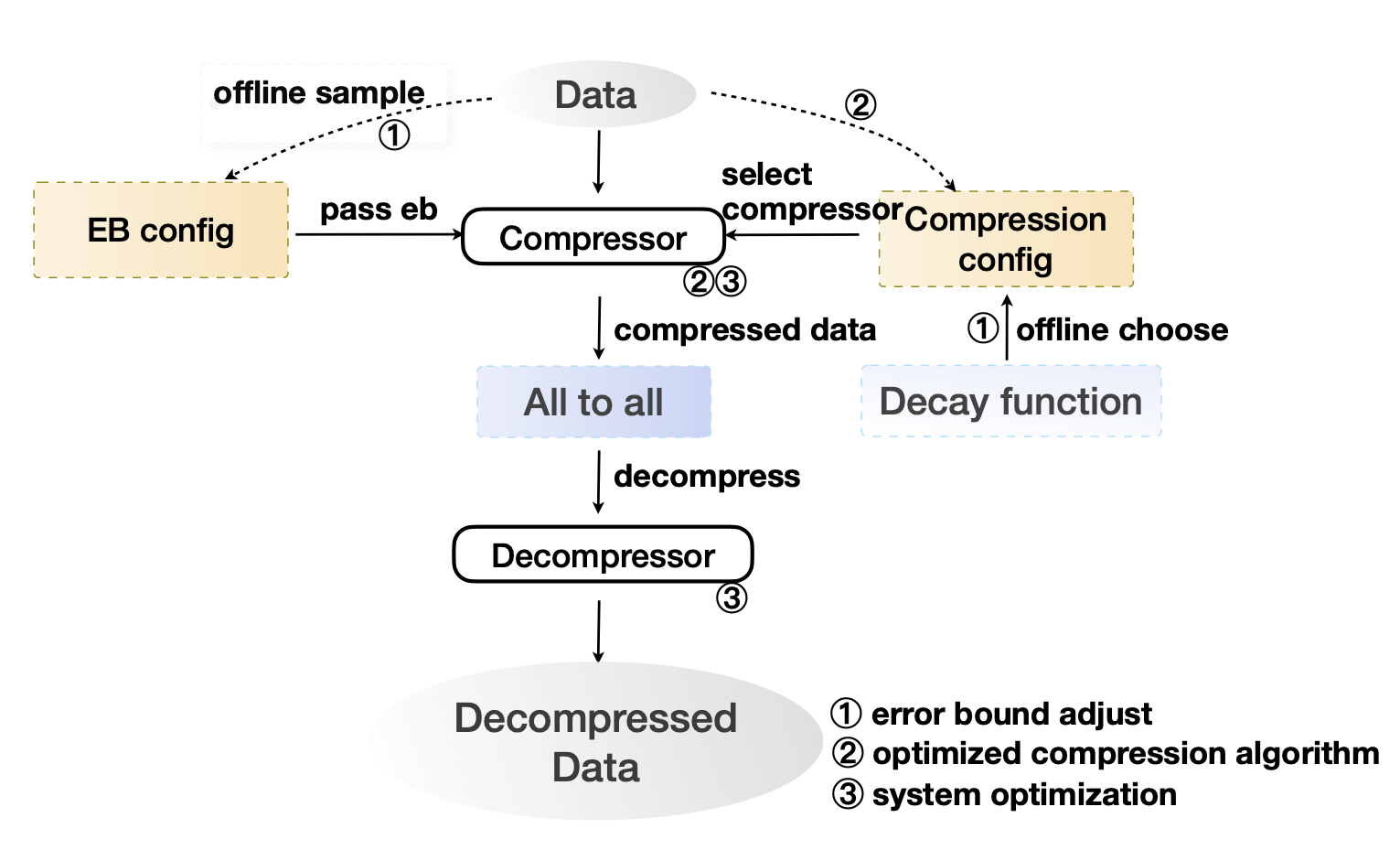}
\caption{Overview of proposed DLRM training framework}
\label{pipeline_overview}
\end{figure}

First, we present the complete framework, showcasing the overview pipeline in Figure \ref{pipeline_overview}. This pipeline can be divided into two main components: an offline analysis process and a training pipeline that incorporates lossy compression.

\textbf{Offline Analysis}. The purpose of this step is to obtain an optimized configuration by sampling and analyzing some iterations from the original training process. There are two tasks involved in offline analysis: Compressor Selection and Embedding Table Classification. In the Compressor Selection task, we evaluate various compressors using sampled data to select the best one for the current system. In the Embedding Table Classification task, we analyze the characteristics of embedding tables using sampled data and classify them according to different error-bound adjustment strategies.

\textbf{Training Pipeline with Compression.} Our proposed training process incorporates lossy compression into all-to-all communications. Unlike fix-rate compression, error-bounded compression does not maintain a consistent compression ratio, making our pipeline distinctive by accommodating variable-size all-to-all communication. The pipeline is organized into four primary stages: \ding{172} Compressing data on each device; \ding{173} Sending metadata through all-to-all communication; \ding{174} Transmitting compressed data via all-to-all communication; and \ding{175} Decompressing data on each device for training. Stages \ding{172} and \ding{175} introduce additional steps for compression and decompression, respectively, where we employ our online error bound adjustment strategy to dynamically tune the error bound. Stage \ding{173} addresses the challenges of executing variable-size all-to-all by managing metadata, including the size of compressed data and compressor specifications.

\subsection{Observation and In-Depth Analysis of Embedding Vector}
\label{sub2:observation}

\label{ob_1} 
\begin{figure}[b]
\centering 
\includegraphics[width=.8\linewidth]{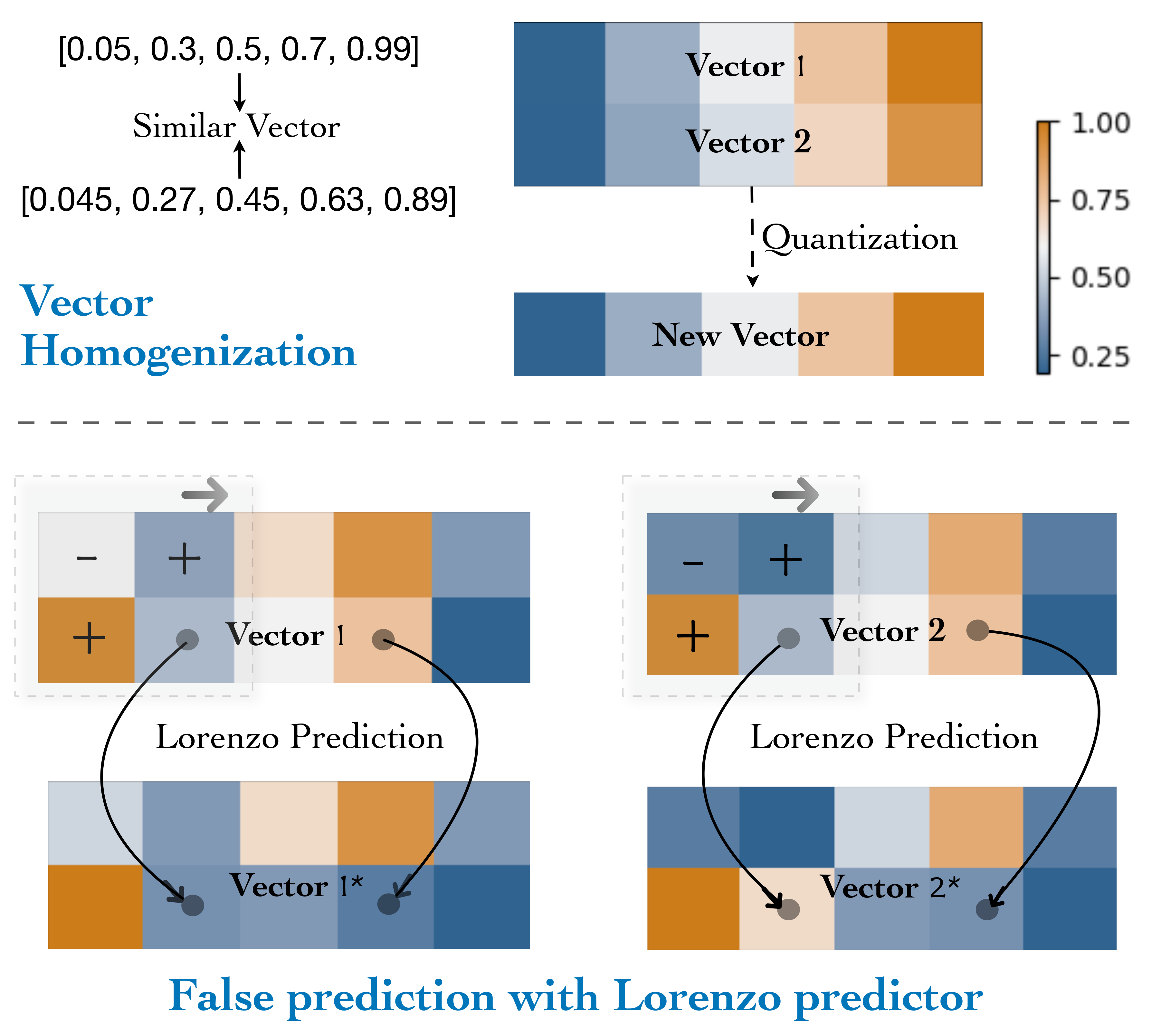} 
\caption{Illustration of observed Vector Homogenization and false prediction with Lorenzo predictor.} 
\label{false_prediction}
\end{figure}

\ding{182} \textbf{False Prediction}. Prediction is a crucial technique in lossy compression algorithms like SZ \cite{sz16, sz17}, leveraging spatial correlations among data points to estimate the value of a point based on its neighbors, as seen with the Lorenzo predictor we mentioned. This approach is effective in many scientific datasets where floating-point numbers represent real-world phenomena, thanks to substantial spatial correlation. However, DLRM embedding vectors are markedly different from scientific data. In a batch of embedding vectors, the spatial correlation is minimal, both within individual vectors and among neighboring ones. It is attributed to the independence of data points across dimensions within an embedding vector and the random order of the vectors. In contrast, the use of prediction can even result in \textit{False Prediction} (illustrated in Figure \ref{false_prediction}), a phenomenon we will elaborate on subsequently.

\label{ob_2} \ding{183} \textbf{Vector Homogenization}. This phenomenon, stemming from quantization and precision loss, significantly impacts data representation. Note that repeated vectors occur not only in original EMB vectors but also increase after quantization, as depicted in Figure \ref{false_prediction}. Within a lossy compression algorithm, two distinct floating-point values within an error-bound range can be considered identical, leading to two EMB vectors being treated as the same if their values at each dimension are sufficiently similar. We term this occurrence \textit{Vector Homogenization}, where similar vectors are transformed into more repetitive ones. Our findings indicate that this phenomenon is more pronounced in certain tables compared to others, attributed to the unique data characteristics of those tables.

\label{ob_3} \ding{184} \textbf{Gaussian Distribution of Data Values}.  In our analysis of the distribution of embedding vectors across different embedding tables, we observe that the distributions tend to vary between Gaussian and uniform, contingent upon the specific table. Embedding tables characterized by significantly unbalanced query frequencies are more inclined to demonstrate a Gaussian distribution. This is attributed to repeated vectors, which result in certain values appearing more frequently, hence deviating the distribution from a uniform to a Gaussian pattern.

\begin{SCtable}
\renewcommand{\arraystretch}{1.2}
\centering\sffamily\footnotesize \caption{Characteristics of their representative EMB tables from Criteo Kaggle dataset.}
\label{table:Data characteristics of some selected EMB tables}
\begin{tabular}{@{} l c c c @{}}
\toprule
EMB Table ID & 1 & 3 & 4 \\ 
\midrule
False Prediction & $\checkmark$ & $\checkmark$ & $\checkmark$ \\ 
\TwoLineTabTitle[l]{Violently Vector}{Homogenization} & $\checkmark$ & $\times$ & $\times$ \\ 
\TwoLineTabTitle[l]{Gaussian}{Distribution} & $\checkmark$ & $\checkmark$ & $\times$ \\ 
\bottomrule
\end{tabular}
\end{SCtable}

\subsection{Adaptive Fine-Grain Error-Bound Adjust Strategy}
\label{sub3:eb_adjust}
Next, we discuss our two-level adaptive strategy for selecting the error bound to ensure the high accuracy of the trained model. As the error bound increases, the compression ratio also increases, but the precision of the decompressed data decreases. There exists a trade-off between reducing communication data size and preserving original information. To address this, we introduce an adaptive strategy for choosing the error bound along two dimensions.

\ding{182} \textbf{Iteration-wise Configuration}. This approach involves gradually decreasing the error bound over iterations, akin to adjusting the learning rate during model training. Two well-recognized insights support this strategy: first, applying different learning rate schedules can yield diverse convergence outcomes for the same model. Second, optimizers do not guarantee an optimal direction for gradient optimization; instead, they aim for a sub-optimal approach, meaning they can tolerate noise on the ideal gradient. Viewing the effects of lossy compression as introducing noise to intermediate data, this noise affects calculations, impacting the gradient during backward propagation. Given that controllable noise does not result in model non-convergence, the key consideration is the acceptable noise amplitude. Similarly to how the learning rate determines the optimization step size, the error bound can be adjusted over time. In the early stages of training, a larger error bound does not hinder convergence. However, as training advances and optimization steps require greater precision, tightening the error bound becomes necessary to limit the noise's impact.

Specifically, we divide the training period into two phases: an \textbf{initial phase}, characterized by rapid loss reduction, and a \textbf{later phase}, where the loss tends to stabilize. During the initial phase, we gradually decrease the error bound {via a predefined decay function (e.g., logarithmic, stepwise) to minimize deviation from the original data, facilitating swift model convergence. In the later phase, we maintain a consistent error bound to ensure the model converges effectively. According to experiment results in Figure \ref{fig:kaggle_decay_func}, it demonstrated that a step-wise (staircase descent) decay function offers the greatest compression benefits while ensuring model convergence. In that case, we select step-wise decay as the default decay function.
\label{section:decay_function}

\begin{figure}
    \begin{subfigure}{\linewidth}
    \centering\includegraphics[width=.8\linewidth, trim={0 5mm 0 0}]{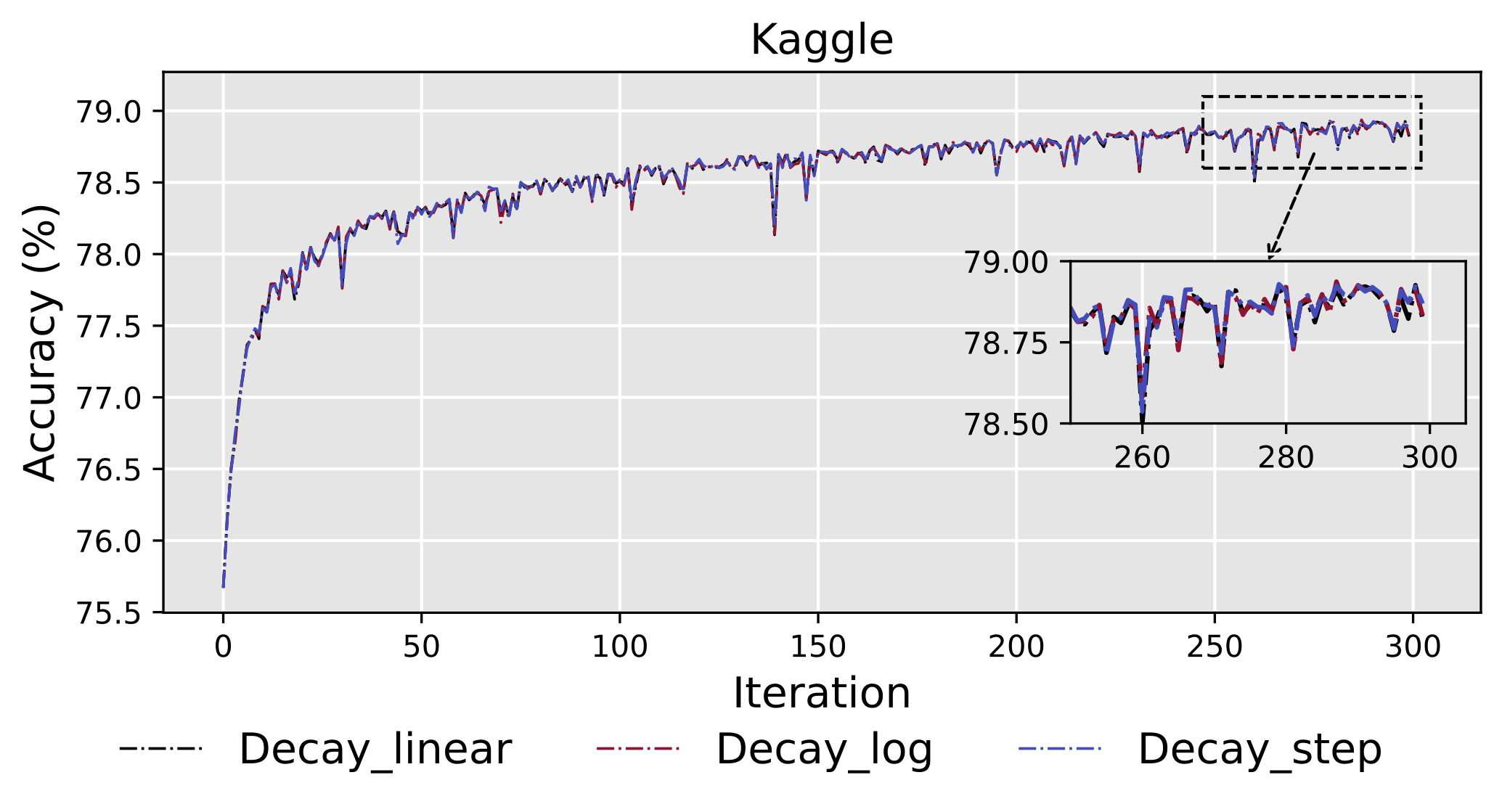}
	\caption{Accuracy}
	\label{kaggle_decay_func_acc}
    \end{subfigure}

    \begin{subfigure}{\linewidth}
    \centering\includegraphics[width=.8\linewidth, trim={0 5mm 0 0}]{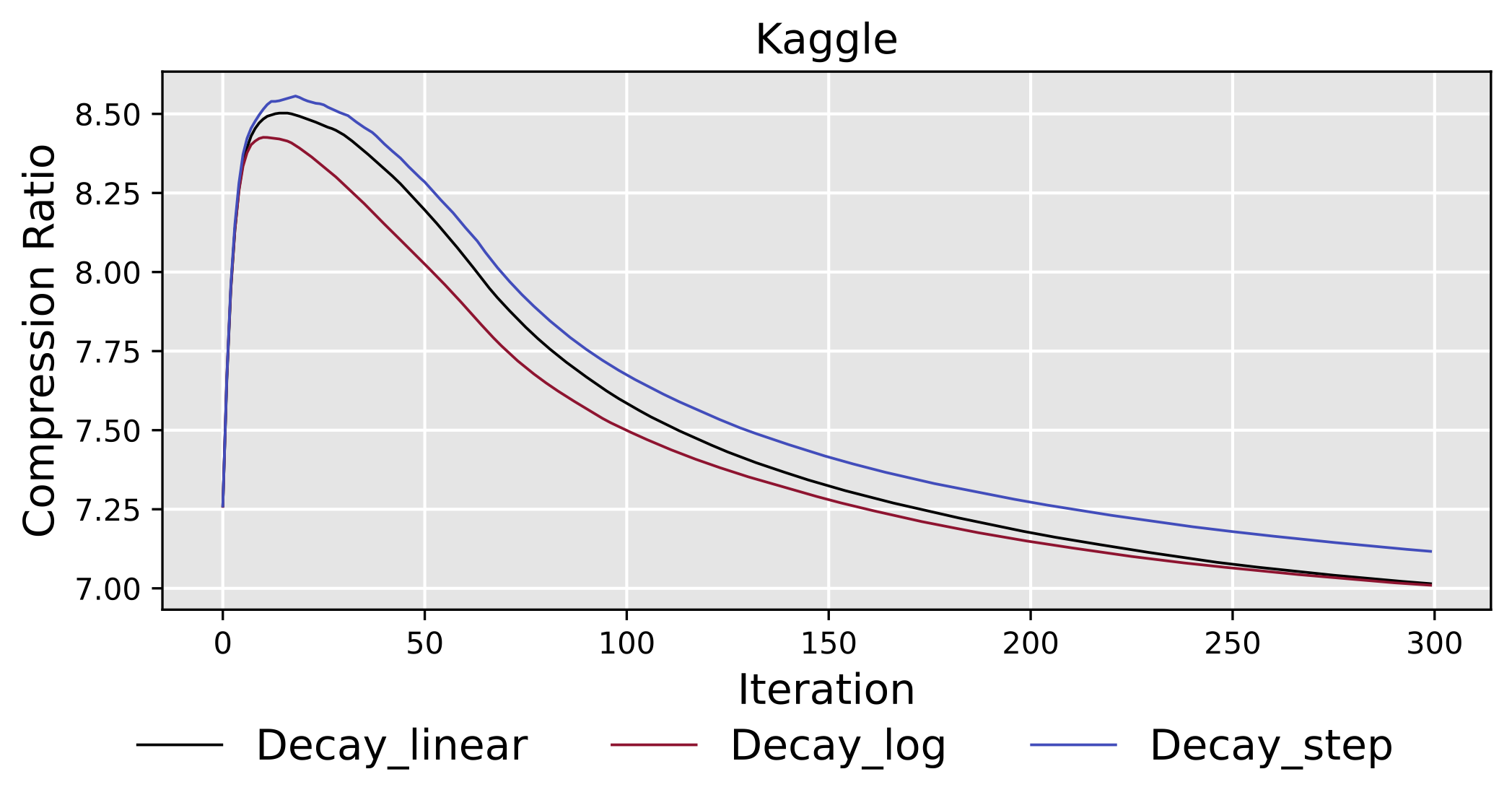}
	\caption{Compression Ratio}
	\label{kaggle_decay_func_avg_cr}
    \end{subfigure}
    
 \caption{Accuracy and CR with different decay functions.}
\label{fig:kaggle_decay_func}
\end{figure}

\ding{183} \textbf{Table-wise Configuration}. Different embedding tables necessitate distinct error bounds, a principle stemming from the inherent properties of the lossy compression algorithm: data quality varies with different characteristics even under identical compression ratios. Embedding vectors within tables symbolize items with diverse semantic meanings. Given the wide variation in embedding table sizes—ranging from fewer than ten to over a million, as depicted in Figure \ref{EMB_size}—the data characteristics among embedding tables significantly diverge.

\begin{figure}[t] 
\centering
\includegraphics[width=.8\linewidth]{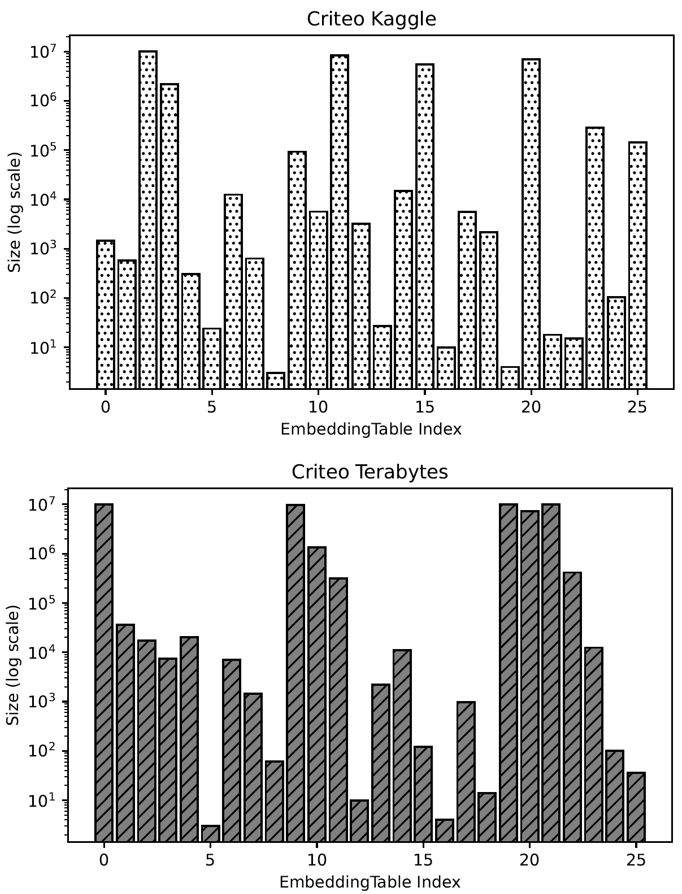}
\caption{EMB table sizes in Criteo Kaggle and Terabytes datasets.}
\label{EMB_size}
\end{figure}

Specifically, the phenomenon of Vector Homogenization is pivotal for setting error bounds tailored to each embedding table, accounting for their unique semantic representations. This necessitates assigning distinct error bounds to ensure uniform quality across tables. Hence, we introduce the Homogenization Index (written in Homo Index in the following text), a metric to assess the quality of embedding tables. This index spans from 0 to 1, where 0 indicates no homogenization and 1 denotes complete vector homogenization into a singular vector. The Homogenization Index is calculated as follows:
\begin{equation}
    \eta = \frac{N_{\text{original}} - N_{\text{compressed}}}{N_{\text{original}}}.
    \label{eq:HOMO INDEX}
\end{equation}
Subsequently, embedding tables are categorized into three groups based on their Homo Index, corresponding to three levels of error bounds: Large, Medium, and Small.

\textbf{Distinguishing Error Bound and Homo Index}. While the error bound for DLRM embedding vectors primarily addresses point-wise error, functioning as a black-box metric, the Homogenization Index (Homo Index) sheds light on the compressed embedding vectors' quality, facilitating adaptive error bound adjustments according to specific requirements. Note that Error Bound and Homo Index serve different purposes and operate on distinct levels: the former at the \textit{point-wise level} and the latter at the \textit{vector-wise level}.

\begin{figure}[htbp]
\begin{lstlisting}[language=Python, basicstyle=\ttfamily\footnotesize, escapeinside={(*@}{@*)}, morekeywords={func, Global}, literate={'}{{\textquotesingle}}1]
Global EBConf = {}
Global LargeEB (*@$\gets$@*) GlobalEB (*@$\times$@*) (*@$\alpha$@*)
Global SmallEB (*@$\gets$@*) GlobalEB (*@$\div$@*) (*@$\beta$@*)
Global MediumEB (*@$\gets$@*) GlobalEB
Global L_EMB_hindex, S_EMB_hindex
Global DecayPhase
Global func DecayFunc

func OfflineAnalysis():
    for t in EMB_Tables:
        sd (*@$\gets$@*) sampleData(t)
        hIndex (*@$\gets$@*) homoIndexCal(sd)
        c (*@$\gets$@*) EMBClassification(hIndex)
        if c is 'large': EBConf[t] (*@$\gets$@*) LargeEB
        if c is 'small': EBConf[t] (*@$\gets$@*) SmallEB
        else EBConf[t] (*@$\gets$@*) MediumEB
        
func OnlineDecay(iter: i):
    for t in EMB_Tables:
        if i (*@$\in$@*) DecayPhase:
            EMBConf[t] (*@$\gets$@*) DecayFunc(i) * EMBConf[t]
func EMBClassification(hindex):
    if hindex > S_EMB_hindex:
        return 'small'
    elif hindex < L_EMB_hindex:
        return 'large'
    else: return 'medium'

\end{lstlisting}
\captionof{lstlisting}{Proposed Error Bound Adjustment Strategy}
\label{alg:adaptiveErrorBound}
\end{figure}

Overall, we detail our proposed adaptive error bound adjustment strategy, as outlined in Algorithm \ref{alg:adaptiveErrorBound}. In pseudo-code, line 1 defines the error-bound parameter to adjust. Lines 2 to 7 define hyper-parameter to adjust the error bound. Homo index calculation in line 11 refers to Equation (\ref{eq:HOMO INDEX}).

\subsection{Optimized Compression Algorithm for DLRM Data}
\label{sub4:comp_algo}
After identifying the optimal error bound for training, the next step involves fine-tuning the compressor within that error bound. Compressors typically aim to balance achieving a high compression ratio with maintaining high compression speed. The ideal compressor for our purposes should satisfy three criteria: \ding{182} Operate on the GPU to avoid data transfers between the device and host since embedding vectors reside on the GPU. \ding{183}  Offer high compression throughput to minimize compression overhead and, consequently, accelerate overall DLRM training. \ding{184} Achieve a high compression ratio to maximize the benefits of reduced data volume during communication. To meet these requirements, we propose an optimized hybrid error-bounded lossy compression algorithm tailored specifically for DLRM data.

Our algorithm comprises two main components: a quantization encoder and a lossless encoder. Initially, the quantization encoder converts floating-point numbers into discrete bins, representing them as integers. These integers are then compressed using a hybrid method that incorporates two types of lossless encoders: LZ encoder \cite{zhang2023gpulz}\cite{lz4} and Entropy encoder such as Huffman encoder \cite{huffman}.

\textbf{Vector-based LZ Encoding.} The phenomenon of \textbf{unbalanced queries}, as illustrated in numerous studies \cite{el-rec}, is pivotal in DLRM training. The imbalance in query frequency implies that recognizing frequently recurring queries can dramatically boost the compression ratio. A distinct feature of repetitive patterns in DLRM applications is the consistency of bytes within an embedding vector for repeated queries, independent of the vector's size. This indicates that the length of the repeating pattern is predetermined and constant. While traditional LZ algorithms are designed to identify repeating patterns of varying lengths, we propose to refine the LZ compression algorithm specifically for DLRM by introducing vector-based LZ compression. This innovation significantly diminishes data volume and amplifies the compression ratio for certain embedding tables through effective pattern recognition.

\textbf{Optimized Entropy Encoding.} Our design of the optimized compression algorithm is informed by two critical observations. The first, observation \ding{184}, underscores the effectiveness of an entropy-based compressor, such as Huffman encoding, given the high entropy typically exhibited by such data. The second, observation \ding{182}, highlights that identical vectors within the same batch might be surrounded by different neighboring vectors, which can lead to divergent predictions for vectors that are initially identical. This phenomenon not only risks misrepresentation and loss of identical embedding vectors but also elevates the data's entropy. An example depicted in Figure \ref{false_prediction} (right part) illustrates how employing a 2x2 Lorenzo predictor \cite{lorenzo} on embedding vectors can transform identical vectors into distinct ones. These insights lead to the conclusion that traditional prediction techniques are ill-suited for our DLRM training-specific compression algorithm.

\textbf{Selection Between Two Encoders.}
During the offline analysis phase, we sample data and evaluate the two encoders to identify the most effective one. Given the complexity of many systems, it is not justifiable to compare compressors solely based on compression ratio or throughput. In our proposed compressor, we utilize a sample-based speed-up approximation to determine the optimal compressor. Equation (\ref{eq:Theoretical Speed-up}) illustrates the method for estimating theoretical speed-up. In this equation, $CR$ denotes the compression ratio, $B$ represents network bandwidth, and $T_c$ and $T_d$ refer to the compression and decompression throughputs, respectively.
\begin{equation}
    \text{\itshape speedup} = \frac{1}{\frac{1}{\text{\itshape CR}} + B \times (\frac{1}{T_{c}} + \frac{1}{T_{d}})}
    \label{eq:Theoretical Speed-up}
\end{equation}

Overall, in our hybrid compression framework, we use this formula to pinpoint the most efficient compressor that maximizes speed-up for the given system configuration. We present the detail in Algorithm \ref{alg:compressorSelection}. In this pseudo code, Line 1 defines the compressor selection parameter. Line 2 defines the hyper-parameter for alternative compressors. The speedup calculation in line 6 refers to Equation (\ref{eq:Theoretical Speed-up}). Although theoretically any compression algorithm could be included in our selection pool, for simplicity and effectiveness, we limit our final design to these two encoders.
\begin{figure}[htbp]
\begin{lstlisting}[language=Python, basicstyle=\ttfamily\footnotesize, escapeinside={(*@}{@*)}, morekeywords={func, Global}]
Global TableCompressorConfig = {}
Global Compressors = {}
func OfflineCompConfig(e):
    for t in EMB_Ta\ttfamilybles:
        d (*@$\gets$@*) sampleEMBData(t)
        speedups (*@$\gets$@*) SpeedUpCompute(d, Compressors)
        maxSup, bestCompressor = Max(speedups)
        TableCompressorConfig[t] (*@$\gets$@*) bestCompressor
\end{lstlisting}
\captionof{lstlisting}{Compressor Selection}
\label{alg:compressorSelection}
\end{figure}

\begin{figure}[hbtp]
\centering
\includegraphics[width=.8\linewidth]{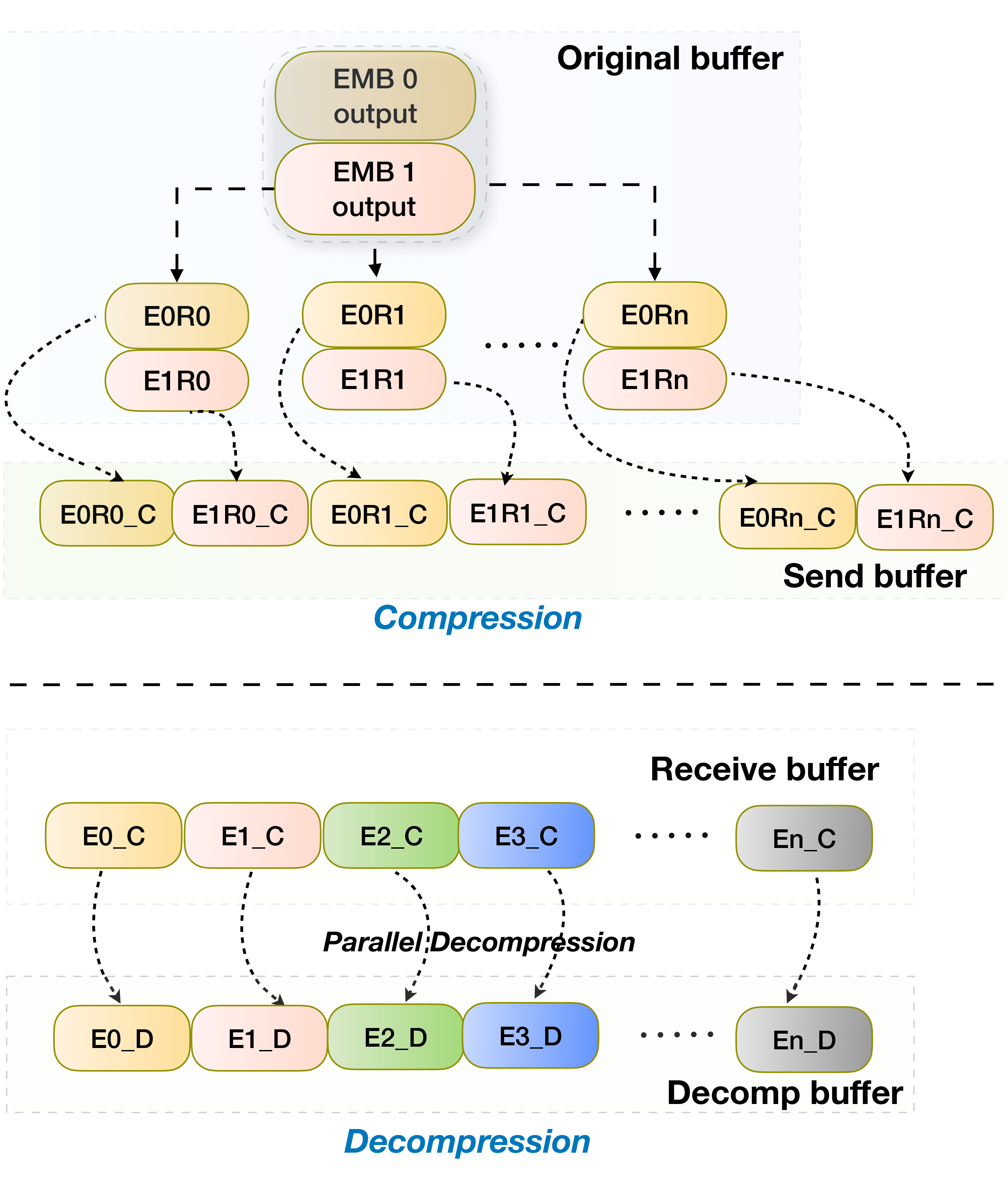}
\caption{Proposed buffer optimization resulting in single-kernel compression (top) and parallel decompression (bottom).} 
\label{fig:Proposed compression and decompression optimization} 
\end{figure}

\subsection{System Implementation and Performance Optimization}
\label{sub5:sysem_opti}
Finally, we detail our system implementation and performance optimization, focusing on fine-tuning the vector-based LZ compressor and GPU compression kernels.

\textbf{Compressor Fine-tuning.} First, we refine our vector-based LZ compression algorithm, distinguishing it from the original LZ approach in two significant ways. \ding{182} \textit{Extended window size}: Traditional LZ algorithms typically opt for a relatively small window size, like 4KB/8KB. However, given the longer repeated patterns observed in DLRM applications, with lengths ranging from 128-256 bytes (considering an embedding vector length of 32/64 with 32-bit floating point data type), we extend the window size to accommodate these patterns. \ding{183} \textit{Fixed pattern length}: To minimize memory access and byte comparison time, we introduce fixed-length pattern matching. In contrast to the standard LZ process, which advances the pointer to the next byte in the absence of a match, our knowledge of the repeated pattern length allows us to leap forward many bytes in search of the next match, thereby avoiding fruitless comparisons and pointer movements. If the initial bytes of two embedding vectors differ, further comparison is unnecessary. This optimization also forces the compressor to match longer patterns rather than shorter patterns. Compared to the current state-of-the-art (SOTA) GPU lossless compressor nvCOMP-LZ4, our approach achieves 2.72$\times$ and 4.88$\times$ higher compression ratios on two datasets, respectively.

\textbf{Buffer Optimization.} Next, we implement multi-threading to minimize memory copy overhead and speed up decompression. Typically, compressors output compressed data to a memory chunk and return a pointer, a versatile but overhead-inducing solution. Given the all-to-all collective communication in DLRM training, where each data chunk must be compressed separately for transmission to each rank, this method introduces unnecessary memory copying since compressed data may not be stored contiguously. Besides, launching kernel multiple times introduces overhead of a lot of kernel launching. To address these, we optimize our compressor that not only introduces one kernel launching but also writes directly to the sending buffer. Figure \ref{fig:Proposed compression and decompression optimization} shows the workflow of compression and decompression in buffer optimization. In the compression process, we involve synchronization and Atomic Add to get the writing offset of each chunk. Furthermore, we leverage multi-threading for compression and decompression, partitioning the compressed data into multiple chunks for simultaneous processing. Although GPU compressors typically achieve high resource utilization, some operations cannot fully leverage GPU resources. Thus, executing multiple decompression kernels in parallel is faster than serial decompression.

\section{Experimental Evaluation}
\label{sec:evaluation}

\subsection{Experimental Setup}

\textbf{Platform, Software, and Dataset.}
Our experimental platforms include a workstation with two NVIDIA A4000 GPUs (16GB memory each) and an HPC cluster comprising 8 GPU nodes. Each node of the cluster is equipped with 256 GB of memory, a 64-core 2.0 GHz 225-watt AMD EPYC 7713 processor, and four NVIDIA A100 GPUs (40GB memory each). Experiments were conducted using PyTorch 2.0.1, CUDA 11.7, and NCCL 2.14.3. We employed the Criteo Ad Kaggle dataset \cite{kaggleDataset2} and the Criteo Terabyte dataset \cite{kaggleDataset} for our experiments. They feature 13 continuous and 26 categorical features, totaling about 45 million samples over 7 days.

\textbf{Baselines.}
To evaluate our design, we compare it against three baselines that focus on reducing communication data volume: \ding{182} The open-source release version of DLRM \cite{dlrm}, serving as the original DLRM training baseline. \ding{183} The low-precision approach, is a straightforward method to reduce communication data volume. Various works \cite{fp8micikevicius2022,fp8shen2023efficient} have demonstrated the feasibility of DLRM training with FP8 data type, making this approach a SOTA solution for reducing communication overhead. \ding{184} We also use nvCOMP \cite{nvcomp}, a software developed by Nvidia that integrates several SOTA lossy and lossless compressors, as a baseline. We compare the compression ratio and throughput between nvCOMP and our optimized compressor.

\definecolor{sizeS}{HTML}{0000FF} %
\definecolor{sizeM}{HTML}{008000} %
\definecolor{sizeL}{HTML}{FF0000} %

\newcommand{\Ssize}{{\textcolor{sizeS}S}}
\newcommand{\Msize}{{\textcolor{sizeM}M}}
\newcommand{\Lsize}{{\textcolor{sizeL}L}}

\newcolumntype{Y}{>{\raggedright\arraybackslash}X}
\begin{table*}[ht!]
\centering
\caption{Classification of EMB tables on test datasets.}
\label{table:EMB_classification}
\centering\sffamily\renewcommand{\arraystretch}{1.2}
\begin{tabularx}{\textwidth}{|l|*{26}{Y|}}
\hline
EMB ID & 0 & 1 & 2 & 3 & 4 & 5 & 6 & 7 & 8 & 9 & 10 & 11 & 12 & 13 & 14 & 15 & 16 & 17 & 18 & 19 & 20 & 21 & 22 & 23 & 24 & 25 \\ \hline
Kaggle & \Msize & \Msize & \Ssize & \Ssize & \Msize & \Msize & \Msize & \Msize & \Lsize & \Ssize & \Msize & \Ssize & \Msize & \Msize & \Msize & \Ssize & \Lsize & \Msize & \Msize & \Lsize & \Ssize & \Lsize & \Lsize & \Ssize & \Lsize & \Ssize \\ \hline
Terabytes & \Ssize & \Msize & \Msize & \Msize & \Msize & \Lsize & \Msize & \Msize & \Lsize & \Ssize & \Ssize & \Msize & \Lsize & \Msize & \Msize & \Lsize & \Lsize & \Lsize & \Lsize & \Ssize & \Ssize & \Ssize & \Ssize & \Msize & \Lsize & \Lsize \\ \hline
\end{tabularx}
\end{table*}

\textbf{EMB Tables Classification.}
We apply our proposed Homogenization Index to classify embedding tables into three categories, as detailed in Table \ref{table:EMB_classification}. These categories correspond to large, medium, and small error bounds, denoted as L, M, and S, respectively. In Table \ref{table:home_idx_kaggle} and Table  \ref{table:home_idx_terabytes}, we select some representative EMB Table to show how Homo Index ranking achieves EMB Table Classification.
\begin{table}[htbp]
\centering\sffamily
\caption{Ranked Homo Index on Criteo Kaggle dataset.}\label{table:home_idx_kaggle}
\scalebox{0.85}{%
\begin{tabular}{@{} cccccc @{}}
\toprule
TAB. ID & EB. & 
\TwoLineTabTitle{\# Ori.}{Patterns} & 
\TwoLineTabTitle{\# Quant.}{Patterns} & 
Batch Size & 
Homo Index 
\\
\midrule
20 & 0.01 & 110 & 68 & 128 & 0.618182 \\
11 & 0.01 & 110 & 69 & 128 & 0.627273 \\
2 & 0.01 & 110 & 73 & 128 & 0.663636 \\
15 & 0.01 & 108 & 76 & 128 & 0.703704 \\
3 & 0.01 & 103 & 86 & 128 & 0.834951 \\
23 & 0.01 & 84 & 77 & 128 & 0.916667 \\
25 & 0.01 & 67 & 63 & 128 & 0.940299 \\
0 & 0.01 & 19 & 19 & 128 & 1 \\
1 & 0.01 & 61 & 61 & 128 & 1 \\
\bottomrule
\end{tabular}
}
\end{table}

\begin{table}[htbp]
\centering\sffamily
\caption{Ranked Homo Index on Criteo Terabytes dataset.}\label{table:home_idx_terabytes}
\scalebox{0.85}{%
\begin{tabular}{@{} cccccc @{}}
\toprule
TAB. ID& EB. & 
\TwoLineTabTitle{\# Ori.}{Patterns} & 
\TwoLineTabTitle{\# Quant.}{Patterns} & 
Batch Size & Homo Index \\
\midrule
0 & 0.005 & 1055 & 484 & 2048 & 0.458768 \\
19 & 0.005 & 1072 & 576 & 2048 & 0.537313 \\
21 & 0.005 & 1042 & 623 & 2048 & 0.597889 \\
9 & 0.005 & 1025 & 621 & 2048 & 0.605854 \\
20 & 0.005 & 937 & 795 & 2048 & 0.848453 \\
1 & 0.005 & 983 & 983 & 2048 & 1 \\
2 & 0.005 & 1302 & 1302 & 2048 & 1 \\
\bottomrule
\end{tabular}
}
\end{table}

\subsection{Evaluation of Error Bound Adjustment Strategy} %

To assess the effects of lossy compression on model accuracy, we evaluate the model prediction accuracy of our training method compared with original DLRM training using both FP32 and FP16 precisions, alongside a SOTA low precision approach employing 8-bit quantization. Current standards deem an accuracy loss within 0.02\% as acceptable in production models \cite{0.02threshold}. 

\begin{figure}[hbtp]
    \begin{subfigure}{\linewidth}
    \centering\includegraphics[trim={-5mm 5mm 2mm 0}, width=\linewidth]{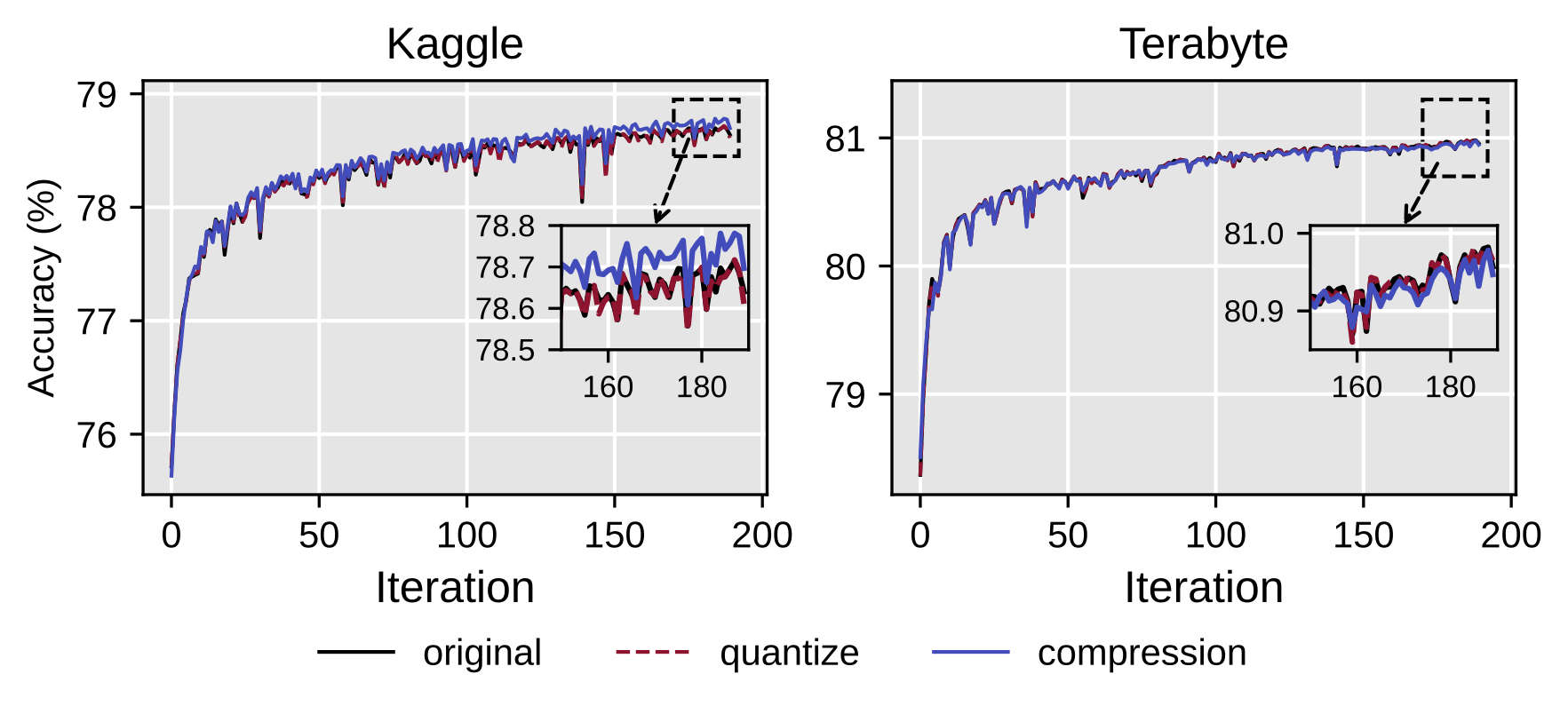}
    \caption{Accuracy}
    \label{kaggle_fp_acc}
    \end{subfigure}
    \begin{subfigure}{\linewidth}
    \centering\includegraphics[trim={0 4mm 0 0}, width=\linewidth]{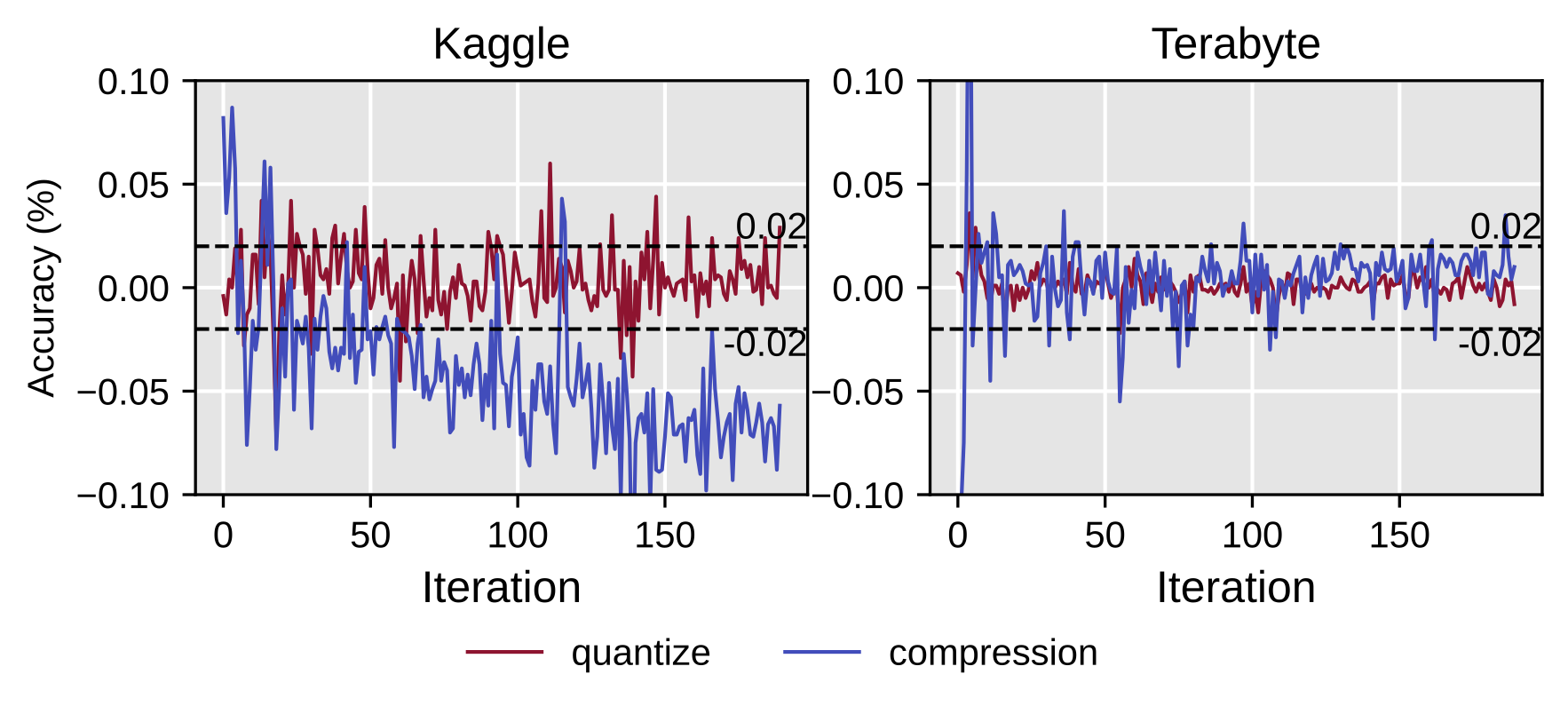}
    \caption{Delta accuracy}
    \label{kaggle_fp_delta}
    \end{subfigure}
     
    \caption{(a) Accuracy and (b) delta accuracy (versus baseline) with different compression methods.}
    \label{fig:Accuracy and delta accuracy with different compression method}
    \vspace{-2mm}
\end{figure}

Figure \ref{kaggle_fp_acc} and figure \ref{kaggle_fp_delta} display a comparison of accuracy convergence curves and accuracy delta curves across these methods, respectively. Through extensive experimentation, we determine and apply a suitable fixed global error bound (i.e., 0.02) for each model to guarantee convergence. On the two datasets we utilized, the average prediction accuracy losses are 0.0031\% and 0.0042\%, respectively.

\textbf{Table-wise Error Bound Adjustment.} Next, to evaluate the effectiveness of our table-wise error bound adjustment strategy, we evaluate the model prediction accuracy and compression ratio throughout the training process, comparing the use of a fixed global error bound against tailored error bounds for different embedding tables. 

As illustrated in Table \ref{table:EMB_classification}, embedding tables are classified into three categories according to their Homogenization Index scores, with assigned error bounds of 0.01, 0.03, and 0.05, respectively. The accuracy evaluation, detailed in Figure \ref{kaggle_global}, reveals that our approach, which applies specific error bounds to different tables rather than a uniform global error bound, maintains the model's accuracy intact. Additionally, this method achieves a higher compression ratio, up to 1.21$\times$ on the Criteo Kaggle dataset, compared to the fixed global error bound strategy.

\begin{figure}[ht]

    \begin{subfigure}{\linewidth}
    \centering\includegraphics[width=\linewidth, trim={0 5mm 0 0}]{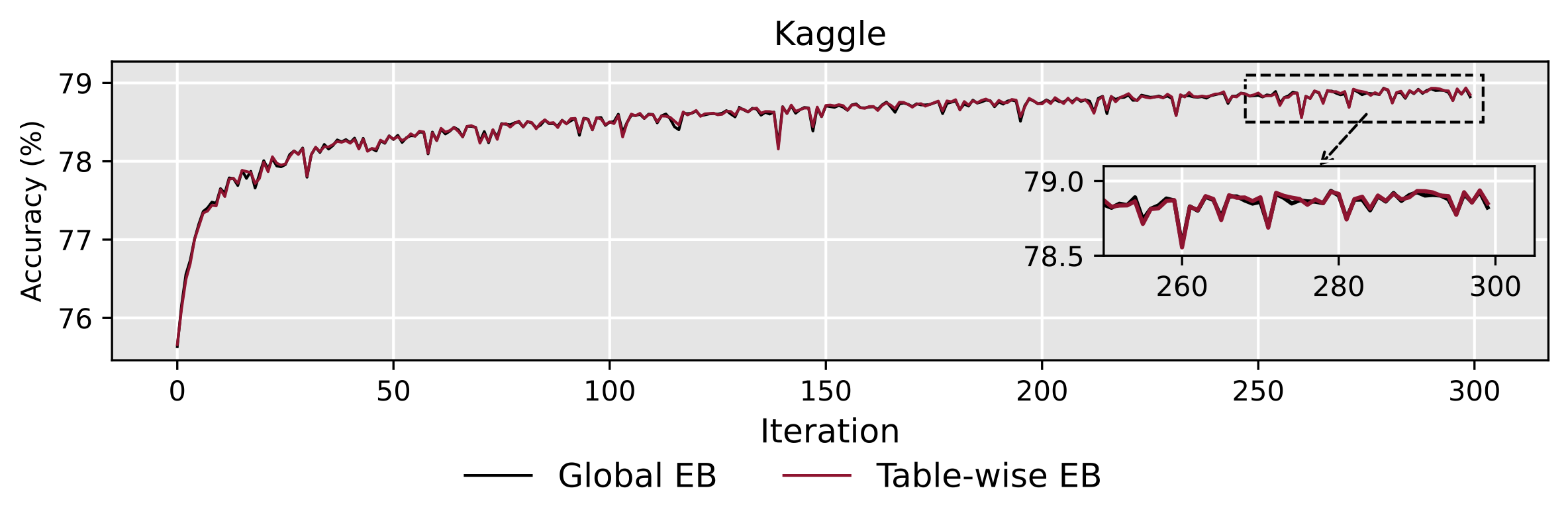}
	\caption{Accuracy}
	\label{kaggle_global}
    \end{subfigure}

    \begin{subfigure}{\linewidth}
    \centering\includegraphics[width=\linewidth, trim={0 5mm 0 0}]{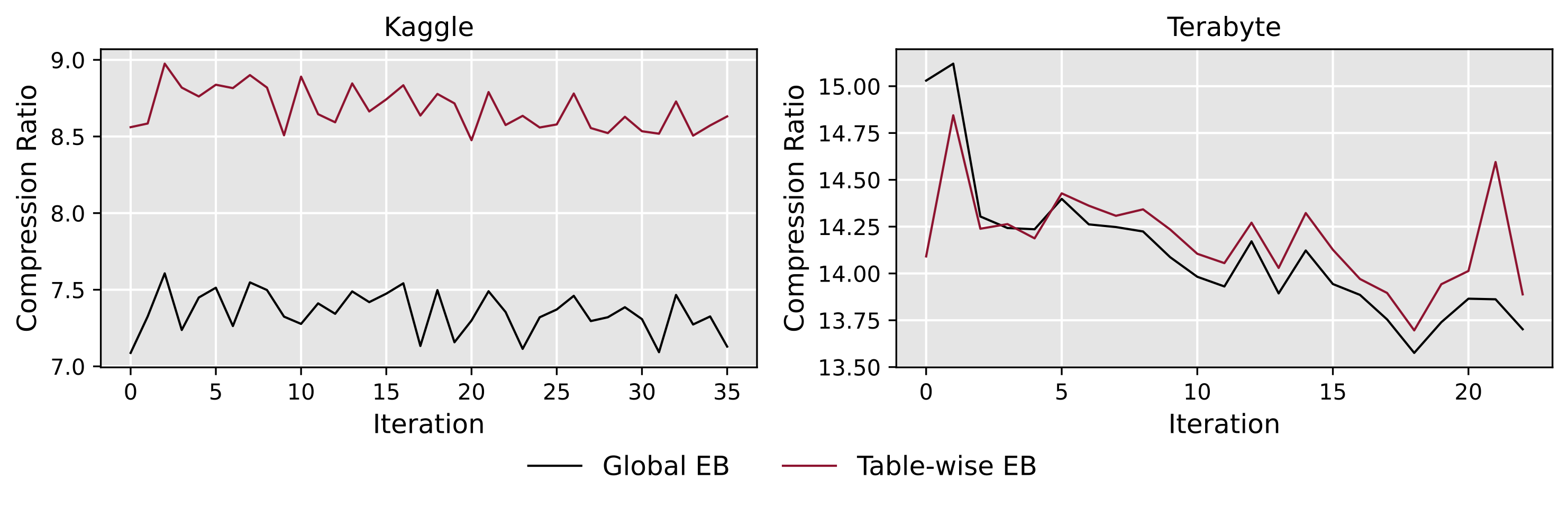}
	\caption{Compression Ratio}
	\label{kaggle_global_cr}
    \end{subfigure}
     
 \caption{Accuracy and compression ratio of our method with proposed table-wise EB configuration strategy on different datasets. e.g. subgraph(b) represents compression ratio on embedding table 0}
\end{figure}

\begin{figure}[ht]

    \begin{subfigure}{\linewidth}
    \centering\includegraphics[width=\linewidth, trim={0 5mm 0 0}]{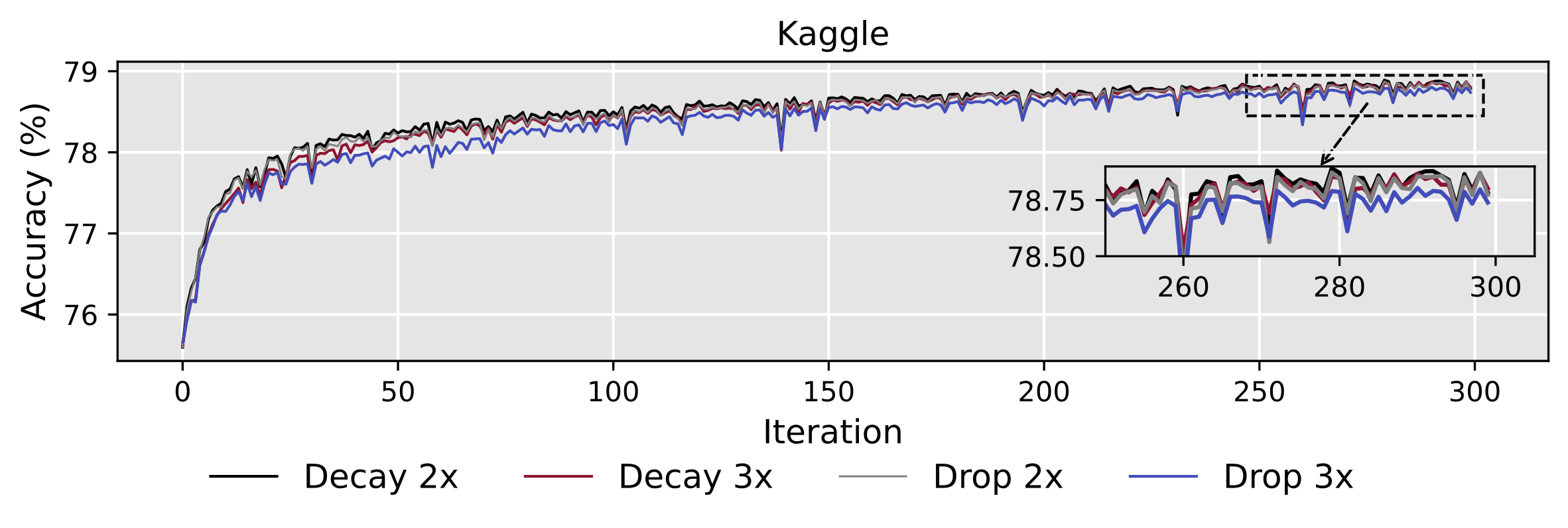}
	\caption{Accuracy}
	\label{kaggle_cons_acc}
    \end{subfigure}

    \begin{subfigure}{\linewidth}
    \centering\includegraphics[width=\linewidth, trim={0 5mm 0 0}]{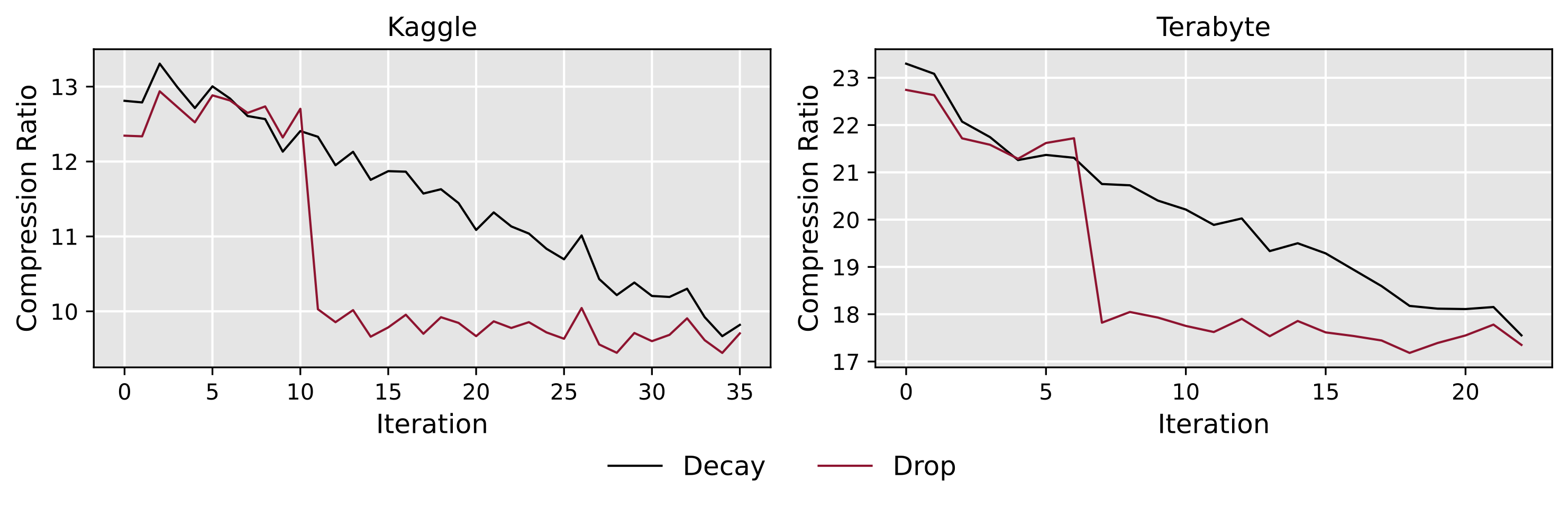}
	\caption{Compression Ratio}
	\label{kaggle_cons_cr}
    \end{subfigure}
     
 \caption{Accuracy and compression ratio of our method with two decay methods on different datasets. Decay$\_$2x represents the error bound decayed from 2 times base error bound to base error bound; Drop$\_$2x represents the error bound dropped from 2 times base error to base error bound at the end of the initial phase. E.g. subgraph (b) represents the compression ratio on embedding table 1.}
 \label{fig:Accuracy and compression ratio with different decay method}
\end{figure}

\textbf{Error Bound Decay Strategy.} Furthermore, to evaluate the effectiveness of our error bound decay strategy, we compare model prediction accuracy and compression ratio throughout the training process using two distinct day approaches: a more aggressive method that abruptly reduces the error bound at a predetermined moment and a gradual approach that decreases the error bound according to a decay function.

As we discussed in \ref{section:decay_function}. This experiment's results of different compressors led us to employ the step-wise function for comparison against the aggressive adjustment approach. Comparative experiments illustrated in Figure \ref{fig:Accuracy and compression ratio with different decay method} reveal how the model training and compression ratios are affected when the error bound is reduced from twice and three times the conservative error bound down to the conservative error bound.

Our analysis indicates that while starting with a larger error bound at the beginning of training and aggressively reducing it can hinder model convergence, a gradual decrease promotes convergence. The comparison of compression ratios clearly shows that, compared to a sharp reduction, our error bound decay strategy allows for starting from a much larger error bound, which is then gradually reduced over time. As a result, this approach yields further 1.09$\times$ and 1.03$\times$ higher compression ratios (i.e., 1.32$\times$ and 1.06$\times$ over the fixed global error bound solution) on Criteo Kaggle and Criteo Terabytes datasets, respectively, delivering more significant benefits.

Based on our evaluations, we choose \textit{{LargeEB: 0.05, MediumEB: 0.03, SmallEB: 0.01, Decay Func: stepwise}} as the optimal error bound configuration for subsequent evaluations.

\begin{figure}[t]
\centering\vspace{-1\baselineskip}
\includegraphics[width=.8\linewidth]{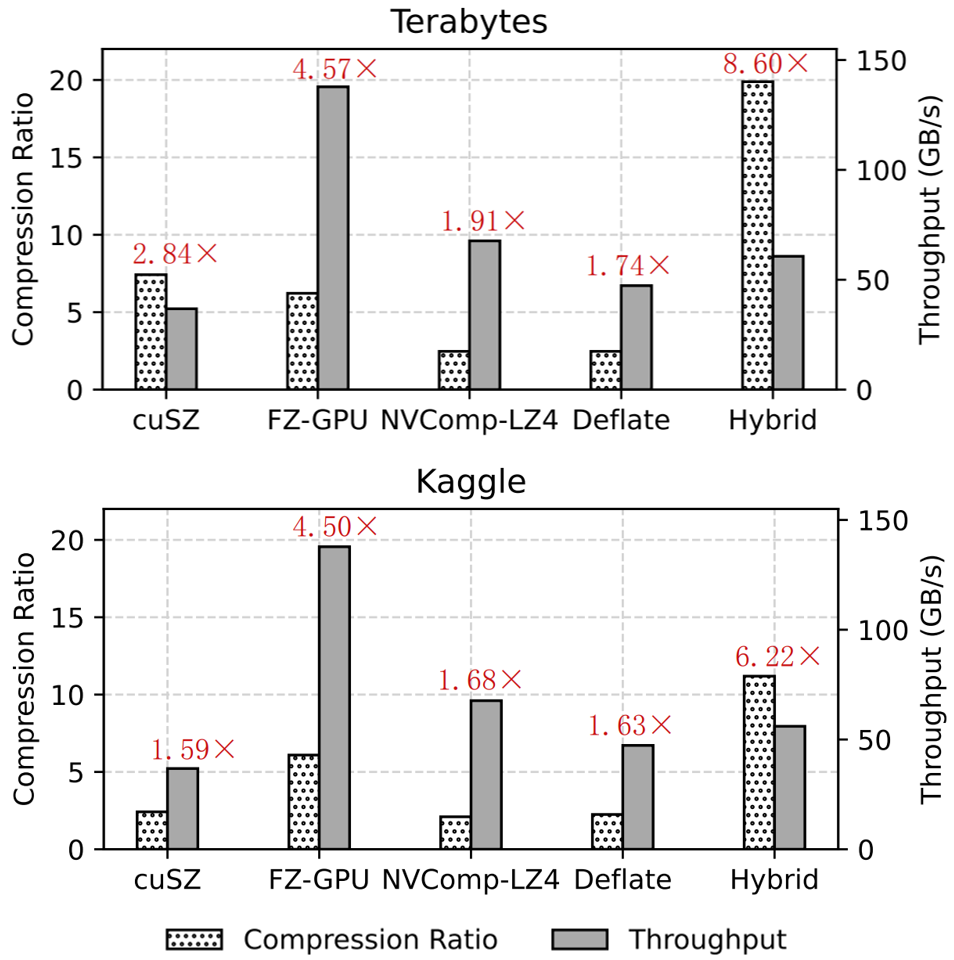}
\caption{Compression ratio, throughput, and communication speedup of different compression methods on different datasets. Batch size = 128 (Kaggle), 2048 (Terabytes).}
\label{kaggle_cr_throughput_and_speedup}
\end{figure}

\subsection{Evaluation on Compression Performance}
In this section, we evaluate the overall compression performance including compression ratio and throughput of different compressors on different DLRM datasets and models. 

\textbf{Overall Compression and Communication Performance.} Figure \ref{kaggle_cr_throughput_and_speedup} illustrates the average compression ratio and throughput during DLRM training. Each sub-graph's left bars depict the average compression ratio of each compressor for a given dataset, while the right bars display each compressor's throughput of compression and decompression. As aforementioned, our hybrid compressor includes two compression algorithms: our vector-based LZ compression algorithm and our optimized entropy compression algorithm.

The results indicate that, for the datasets utilized in DLRM training, our hybrid compressor outperforms other compressors in terms of compression ratio and achieves exceptionally high throughput. It reaches an overall 11.2$\times$ and 19.9$\times$ compression ratio on Criteo Kaggle and Criteo Terabytes, respectively. Our two proposed compressors, vector-based LZ can reach 40.5GB/s in compression, and 205.4GB/s in decompression, while the optimized entropy compressor can reach 78.4GB/s in compression, and 38.9GB/s in decompression. Compared to the SOTA lossless compressors nvCOMP LZ4 our compressor achieves a compression ratio 5.3$\times$ and 8.1$\times$ higher on two datasets respectively. The other SOTA nvCOMP Deflate achieves a similar compression ratio to nvCOMP LZ4 but compression and decompression throughput are 30.1GB/s and 109.7GB/s. The SOTA lossy compressor FZ-GPU \cite{zhang2023fz} has the highest throughput which is over 136GB/s in both compression and decompression, since it relies on a very fast encoder (i.e., bitshuffle and sparse encoding) \cite{zhang2023fz}. However, its compression ratio is significantly lower than our hybrid compressor attained, leading to a greater overall speedup in end-to-end communication.

As shown in Figure \ref{kaggle_cr_throughput_and_speedup}, our proposed hybrid compressor achieves a 6.22$\times$ and 8.6$\times$ speedup for two different datasets in all-to-all communication, surpassing all other approaches when all-to-all communication throughput is 4GB/s.

\textbf{DLRM End-to-end Performance Speed-up} Figure \ref{end_to_end_speedup} shows the breakdown of DLRM training with lossy compression on our cluster with 32 A100 GPUs.  Our compression accelerates all-to-all communication in forward propagation which takes 31.3\% proportion of the whole training time. On the Criteo Kaggle dataset, our proposed compressor achieves 6.22$\times$ overall speed-up in communication and 1.30$\times$ in end-to-end training, by reducing all-to-all in forward propagation cost to 5.03\%. On the Criteo Terabytes dataset, our compressor achieves 8.6$\times$ and 1.38$\times$ in all-to-all communication and end-to-end training, respectively. 

\begin{figure}[ht]
\centering
\includegraphics[width=0.49\textwidth]{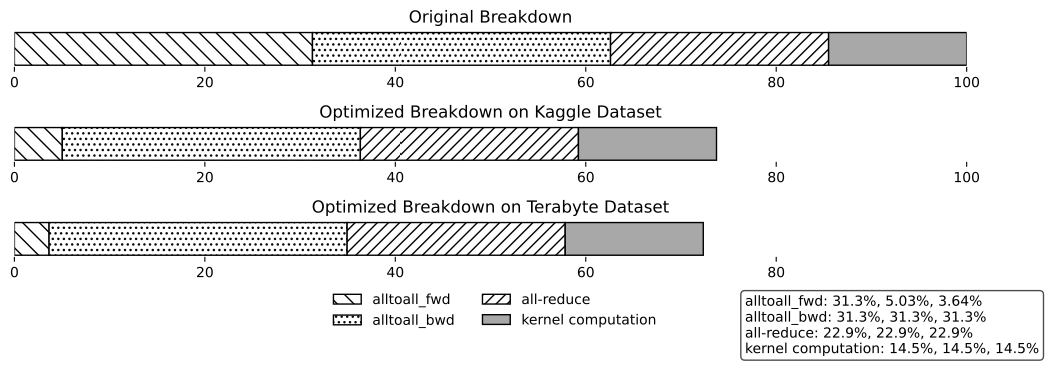}
\caption{Breakdown of optimized end-to-end DLRM training time on different DLRM datasets.}
\label{end_to_end_speedup}
\end{figure}

\begin{table*}
\centering\sffamily\scriptsize
\caption{Compression ratio of different compressors on the two datasets (left: Criteo Kaggle, right: Terabytes). Bolded numbers indicate the highest compression ratios among compressors.}

\color{gray}
    \begin{tabular}{@{} >{\color{black}}l *{7}{c} @{}}
  \rotatebox{90}{\TwoLineTabTitle[l]{EMB}{Table ID}}                       &
  \rotatebox{90}{cuSZ}                                                     &
  \rotatebox{90}{FZ-GPU}                                                   &
  \rotatebox{90}{\color{black}\ThreeLineTabTitle[l]{Ours-}{Vector}{GPULZ}} &
  \rotatebox{90}{\color{black}\TwoLineTabTitle[l]{Ours-}{Huffman}}         &
  \rotatebox{90}{\ThreeLineTabTitle[l]{Baseline}{NVComp-}{LZ4}}            &
  \rotatebox{90}{\TwoLineTabTitle[l]{Baseline}{Deflate}}                   &
  \rotatebox{90}{\ThreeLineTabTitle[l]{Huffman+}{GPULZ}{hybrid}}
  \\
  \toprule
  0                                                                        & \phantom{00}2.40 / \phantom{00}6.40                    & \phantom{00}5.84 / \phantom{00}5.14                    & \phantom{00}7.58 / \TabHighlightData{\phantom{0}14.94}                       & \TabHighlightData{\phantom{00}8.74} / \NotTabHighlightData{\phantom{0}10.94} & \phantom{00}5.87 / \phantom{00}1.62                    & \phantom{00}6.32 / \phantom{00}1.67                    & \phantom{00}8.74 / \phantom{0}14.94                             \\
  1                                                                        & \phantom{00}2.40 / \phantom{00}9.90                    & \phantom{00}6.45 / \phantom{00}7.70                    & \phantom{00}3.96 / \phantom{00}5.93                                          & \TabHighlightData{\phantom{0}11.20} / \phantom{0}21.44                       & \phantom{00}1.90 / \phantom{00}1.30                    & \phantom{00}2.05 / \phantom{00}1.34                    & \phantom{0}11.20 / \phantom{0}21.44                             \\
  2                                                                        & \phantom{00}2.34 / \phantom{00}8.67                    & \phantom{00}5.37 / \phantom{00}7.28                    & \phantom{00}4.46 / \phantom{00}4.47                                          & \TabHighlightData{\phantom{00}9.44 / \phantom{0}17.18}                       & \phantom{00}1.11 / \phantom{00}1.11                    & \phantom{00}1.17 / \phantom{00}1.18                    & \phantom{00}9.44 / \phantom{0}17.18                             \\
  3                                                                        & \phantom{00}2.30 / \phantom{00}6.53                    & \phantom{00}5.33 / \phantom{00}5.87                    & \phantom{00}4.01 / \TabHighlightData{\phantom{0}16.44}                       & \TabHighlightData{\phantom{00}8.78} / \NotTabHighlightData{\phantom{0}10.13} & \phantom{00}1.19 / \phantom{00}3.69                    & \phantom{00}1.27 / \phantom{00}3.44                    & \phantom{00}8.78 / \phantom{0}16.44                             \\
  4                                                                        & \phantom{00}2.53 / \phantom{00}6.64                    & \phantom{00}6.22 / \phantom{00}5.98                    & \TabHighlightData{\phantom{0}12.96} / \NotTabHighlightData{\phantom{0}10.49} & \phantom{00}8.68 / \TabHighlightData{\phantom{0}10.65}                       & \phantom{0}10.23 / \phantom{00}2.46                    & \phantom{0}10.65 / \phantom{00}2.42                    & \phantom{0}12.96 / \phantom{0}10.65                             \\
  5                                                                        & \phantom{00}2.39 / \phantom{0}12.34                    & \phantom{00}6.05 / \phantom{00}9.87                    & \TabHighlightData{\phantom{00}9.90} / 915.47                                 & \phantom{00}9.81 / \phantom{0}11.03                                          & \phantom{0}12.86 / \phantom{0}67.84                    & \phantom{0}13.71 / \phantom{0}41.44                    & \phantom{00}9.90 / 915.47                                       \\
  6                                                                        & \phantom{00}2.50 / \phantom{00}8.38                    & \phantom{00}6.99 / \phantom{00}7.15                    & \phantom{00}3.88 / \phantom{00}4.87                                          & \TabHighlightData{\phantom{0}14.03} / \phantom{0}16.02                       & \phantom{00}1.08 / \phantom{00}1.20                    & \phantom{00}1.16 / \phantom{00}1.25                    & \phantom{0}14.03 / \phantom{0}16.02                             \\
  7                                                                        & \phantom{00}2.45 / \phantom{00}6.77                    & \phantom{00}6.13 / \phantom{00}6.07                    & \TabHighlightData{\phantom{00}9.96} / \NotTabHighlightData{\phantom{00}9.38} & \phantom{00}8.67 / \TabHighlightData{\phantom{0}11.30}                       & \phantom{00}8.16 / \phantom{00}2.17                    & \phantom{00}8.61 / \phantom{00}2.07                    & \phantom{00}9.96 / \phantom{0}11.30                             \\
  8                                                                        & \phantom{00}2.81 / \phantom{00}8.33                    & \phantom{00}8.92 / \phantom{00}6.84                    & \TabHighlightData{\phantom{0}41.92} / 124.77                                 & \phantom{0}10.05 / \phantom{0}13.46                                          & \phantom{0}37.55 / \phantom{0}16.81                    & \phantom{0}34.06 / \phantom{0}11.68                    & \phantom{0}41.92 / 124.77                                       \\
  9                                                                        & \phantom{00}2.23 / \phantom{00}6.22                    & \phantom{00}4.93 / \phantom{00}5.09                    & \phantom{00}4.07 / \TabHighlightData{\phantom{0}11.77}                       & \TabHighlightData{\phantom{00}7.55} / \NotTabHighlightData{\phantom{0}10.31} & \phantom{00}1.32 / \phantom{00}1.63                    & \phantom{00}1.42 / \phantom{00}1.68                    & \phantom{00}7.55 / \phantom{0}11.77                             \\
  10                                                                       & \phantom{00}2.48 / \phantom{00}6.07                    & \phantom{00}6.84 / \phantom{00}5.18                    & \phantom{00}3.87 / \phantom{00}9.34                                          & \TabHighlightData{\phantom{0}13.28} / \phantom{00}9.51                       & \phantom{00}1.15 / \phantom{00}1.98                    & \phantom{00}1.24 / \phantom{00}1.97                    & \phantom{0}13.28 / \phantom{00}9.51                             \\
  11                                                                       & \phantom{00}2.34 / \phantom{00}9.14                    & \phantom{00}5.30 / \phantom{00}6.82                    & \phantom{00}4.55 / \phantom{00}8.03                                          & \TabHighlightData{\phantom{00}9.28} / \phantom{0}18.69                       & \phantom{00}1.12 / \phantom{00}1.31                    & \phantom{00}1.19 / \phantom{00}1.38                    & \phantom{00}9.28 / \phantom{0}18.69                             \\
  12                                                                       & \phantom{00}2.48 / \phantom{00}7.32                    & \phantom{00}6.73 / \phantom{00}6.37                    & \phantom{00}3.87 /  \TabHighlightData{188.52}                                & \TabHighlightData{\phantom{0}13.47} / \NotTabHighlightData{\phantom{0}11.50} & \phantom{00}1.19 / \phantom{0}20.84                    & \phantom{00}1.28 / \phantom{0}13.59                    & \phantom{0}13.47 / 188.52                                       \\
  13                                                                       & \phantom{00}2.38 / \phantom{00}7.40                    & \phantom{00}5.75 / \phantom{00}6.33                    & \TabHighlightData{\phantom{00}9.74} / \phantom{0}25.54                       & \phantom{00}9.51 / \phantom{0}10.67                                          & \phantom{00}9.68 / \phantom{00}5.92                    & \phantom{0}10.29 / \phantom{00}5.83                    & \phantom{00}9.74 / \phantom{0}25.54                             \\
  14                                                                       & \phantom{00}2.56 / \phantom{00}8.75                    & \phantom{00}7.33 / \phantom{00}7.33                    & \phantom{00}3.88 / \phantom{00}6.48                                          & \TabHighlightData{\phantom{0}15.27} / \phantom{0}17.94                       & \phantom{00}1.14 / \phantom{00}1.55                    & \phantom{00}1.22 / \phantom{00}1.54                    & \phantom{0}15.27 / \phantom{0}17.94                             \\
  15                                                                       & \phantom{00}2.35 / \phantom{00}7.83                    & \phantom{00}5.48 / \phantom{00}6.51                    & \phantom{00}4.23 / \TabHighlightData{\phantom{0}70.26}                       & \TabHighlightData{\phantom{00}9.70} / \NotTabHighlightData{\phantom{0}12.16} & \phantom{00}1.14 / \phantom{0}12.76                    & \phantom{00}1.20 / \phantom{00}9.99                    & \phantom{00}9.70 / \phantom{0}70.26                             \\
  16                                                                       & \phantom{00}2.41 / \phantom{00}7.52                    & \phantom{00}6.38 / \phantom{00}6.35                    & \phantom{00}8.09 / \TabHighlightData{338.78}                                 & \TabHighlightData{\phantom{0}10.49} / \NotTabHighlightData{\phantom{0}11.26} & \phantom{0}10.80 / \phantom{0}36.04                    & \phantom{0}11.53 / \phantom{0}25.06                    & \phantom{0}10.49 / 338.78                                       \\
  17                                                                       & \phantom{00}2.47 / \phantom{00}8.08                    & \phantom{00}6.66 / \phantom{00}6.47                    & \phantom{00}3.88 / \TabHighlightData{\phantom{0}41.66}                       & \TabHighlightData{\phantom{0}12.77} / \NotTabHighlightData{\phantom{0}12.44} & \phantom{00}1.34 / \phantom{00}8.43                    & \phantom{00}1.44 / \phantom{00}7.21                    & \phantom{0}12.77 / \phantom{0}41.66                             \\
  18                                                                       & \phantom{00}2.44 / \phantom{00}7.81                    & \phantom{00}6.25 / \phantom{00}6.49                    & \phantom{00}8.31 / \TabHighlightData{136.59}                                 & \TabHighlightData{\phantom{00}9.40} / \NotTabHighlightData{\phantom{0}11.67} & \phantom{00}4.63 / \phantom{0}18.32                    & \phantom{00}4.94 / \phantom{0}12.53                    & \phantom{00}9.40 / 136.59                                       \\
  19                                                                       & \phantom{00}2.46 / \phantom{00}6.35                    & \phantom{00}6.51 / \phantom{00}5.22                    & \TabHighlightData{\phantom{0}13.72} / \phantom{0}13.72                       & \phantom{0}10.62 / \phantom{0}11.05                                          & \phantom{0}18.62 / \phantom{00}1.61                    & \phantom{0}19.93 / \phantom{00}1.66                    & \phantom{0}13.72 / \phantom{0}13.72                             \\
  20                                                                       & \phantom{00}2.37 / \phantom{00}6.00                    & \phantom{00}5.42 / \phantom{00}5.18                    & \phantom{00}4.51 / \phantom{00}9.00                                          & \TabHighlightData{\phantom{00}9.93} / \phantom{00}9.58                       & \phantom{00}1.13/ \phantom{00}1.69                     & \phantom{00}1.20 / \phantom{00}1.73                    & \phantom{00}9.93 / \phantom{00}9.58                             \\
  21                                                                       & \phantom{00}2.66 / \phantom{00}6.31                    & \phantom{00}6.68 / \phantom{00}5.11                    & \TabHighlightData{\phantom{0}21.50} / \phantom{0}11.81                       & \phantom{0}10.87 / \phantom{0}10.67                                          & \phantom{0}19.23 / \phantom{00}1.62                    & \phantom{0}19.13 / \phantom{00}1.67                    & \phantom{0}21.50 / \phantom{0}11.81                             \\
  22                                                                       & \phantom{00}2.46 / \phantom{00}6.24                    & \phantom{00}6.48 / \phantom{00}5.15                    & \TabHighlightData{\phantom{00}8.40} / \phantom{0}12.59                       & \phantom{0}11.81 / \phantom{00}9.15                                          & \phantom{0}10.40 / \phantom{00}2.73                    & \phantom{0}11.11 / \phantom{00}2.78                    & \phantom{00}8.40 / \phantom{0}12.59                             \\
  23                                                                       & \phantom{00}2.21 / \phantom{00}8.19                    & \phantom{00}5.08 / \phantom{00}7.03                    & \phantom{00}3.99 / \phantom{00}4.75                                          & \TabHighlightData{\phantom{0}12.59 / \phantom{0}15.25}                       & \phantom{00}1.43 / \phantom{00}1.18                    & \phantom{00}1.53 / \phantom{00}1.28                    & \phantom{0}12.59 / \phantom{0}15.25                             \\
  24                                                                       & \phantom{00}2.49 / \phantom{00}7.78                    & \phantom{00}6.95 / \phantom{00}6.58                    & \phantom{00}7.21 / \TabHighlightData{110.97}                                 & \TabHighlightData{\phantom{0}12.03} / \NotTabHighlightData{\phantom{0}12.63} & \phantom{00}7.32 / \phantom{0}16.30                    & \phantom{00}7.87 / \phantom{0}11.67                    & \phantom{0}12.03 / 110.97                                       \\
  25                                                                       & \phantom{00}2.25 / \phantom{00}7.59                    & \phantom{00}5.09 / \phantom{00}6.46                    & \phantom{00}5.27 / \TabHighlightData{\phantom{0}80.88}                       & \TabHighlightData{\phantom{0}12.08} / \NotTabHighlightData{\phantom{0}12.69} & \phantom{00}2.09 / \phantom{0}13.09                    & \phantom{00}2.23 / \phantom{00}9.61                    & \phantom{0}12.08 / \phantom{0}80.88                             \\
  \midrule
  \color{RoyalBlue}{avg.}                                                  & \color{RoyalBlue}{\phantom{00}2.42 / \phantom{00}7.42} & \color{RoyalBlue}{\phantom{00}6.09 / \phantom{00}6.22} & \color{RoyalBlue}{\phantom{00}5.73 / \phantom{0}12.50}                       & \color{RoyalBlue}{\phantom{0}10.45 / \phantom{0}12.06}                       & \color{RoyalBlue}{\phantom{00}2.10 / \phantom{00}2.47} & \color{RoyalBlue}{\phantom{00}2.25 / \phantom{00}2.47} & \TabHighlightData{\textbf{\phantom{0}11.19 / \phantom{0}19.89}} \\
  \bottomrule
\end{tabular}%

\label{table:comp_performance}
\end{table*}

\textbf{Compression Performance across Embedding Tables.} 
Furthermore, we outline the compression ratios our compressor achieved across various embedding tables and delve into the data characteristics that enable our compressor to secure maximum compression benefits.
Table \ref{table:comp_performance} presents the compression ratios for various compressors across different embedding tables. We can draw three key observations from this result. Firstly, the compression ratios for all compressors vary significantly across embedding tables, underscoring the importance of choosing a compressor that's well-suited for each specific table. Secondly, our optimized vector-based LZ algorithm excels with certain embedding tables, while its performance on others is less impressive. Lastly, the performance trends of our optimized entropy-based compressor and the vector-based LZ algorithm appear to be in stark contrast.

We employ data sampling to shed light on the substantial variance in compression ratios among different embedding tables. Figure \ref{EMB_data_feature} illustrates the matched pattern number and data distribution for two representative tables from the Terabytes dataset. The data histogram reveals that \textit{EMB Table 1} exhibits a highly concentrated Gaussian distribution, whereas \textit{EMB Table 5} displays a broad dispersion of data with similar frequencies. This distinction in data entropy explains the higher compression ratio achieved with the Huffman encoder for \textit{EMB Table 1}. Due to the limited unique embedding vectors in \textit{EMB Table 5}, the likelihood of LZ encoder matching patterns is significantly high, resulting in a superior compression ratio. The marked difference in the number of matched patterns elucidates why the LZ encoder outperforms embedding tables like \textit{EMB Table 5}.

\begin{figure}[ht]
    \begin{subfigure}[b]{\linewidth}
    \centering 
    \includegraphics[width=0.7\textwidth, trim={0 5mm 0 0}]{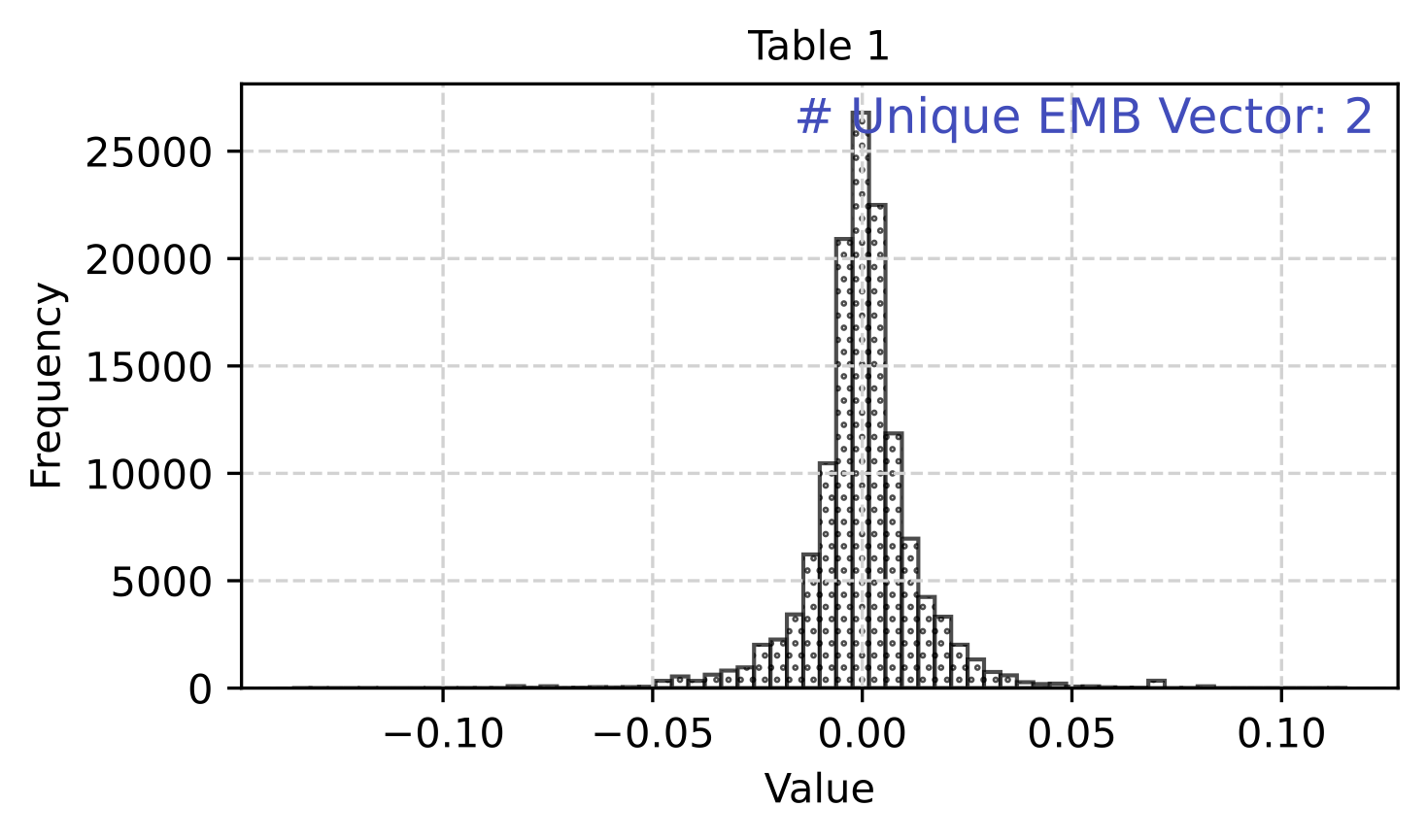}
	\caption{Sampled Batch of EMB Table 1}
	\label{hist_t1}
    \end{subfigure}

    \begin{subfigure}[b]{\linewidth}
    \centering 
    \includegraphics[width=0.7\textwidth, trim={0 5mm 0 0}]{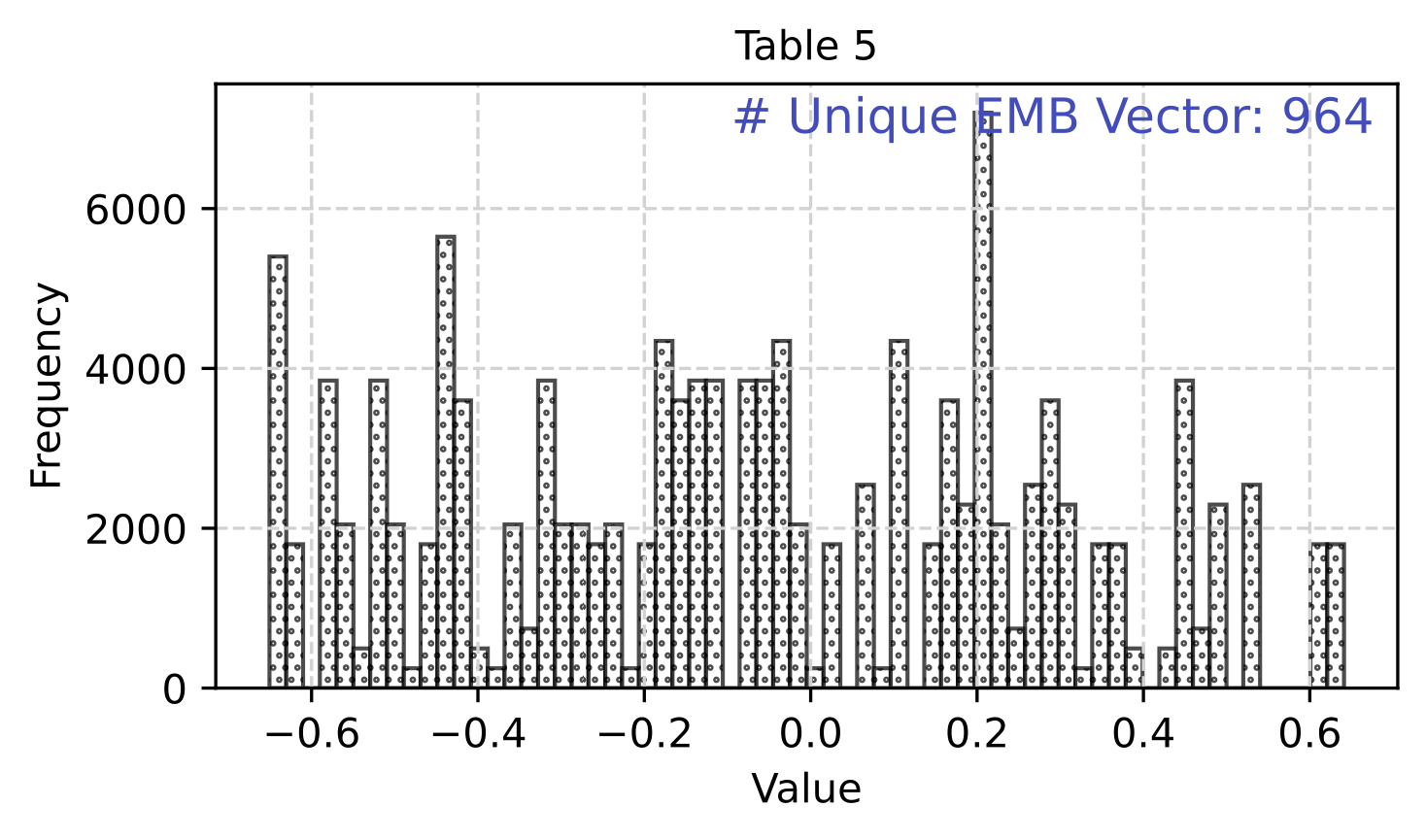}
	\caption{Sampled Batch of EMB Table 5}
	\label{hist_t5}
    \end{subfigure}
    
\caption{Data features of two representative EMB tables.}
\label{EMB_data_feature}
\end{figure}

\begin{figure*}[t]
\centering
\includegraphics[width=.9\textwidth]{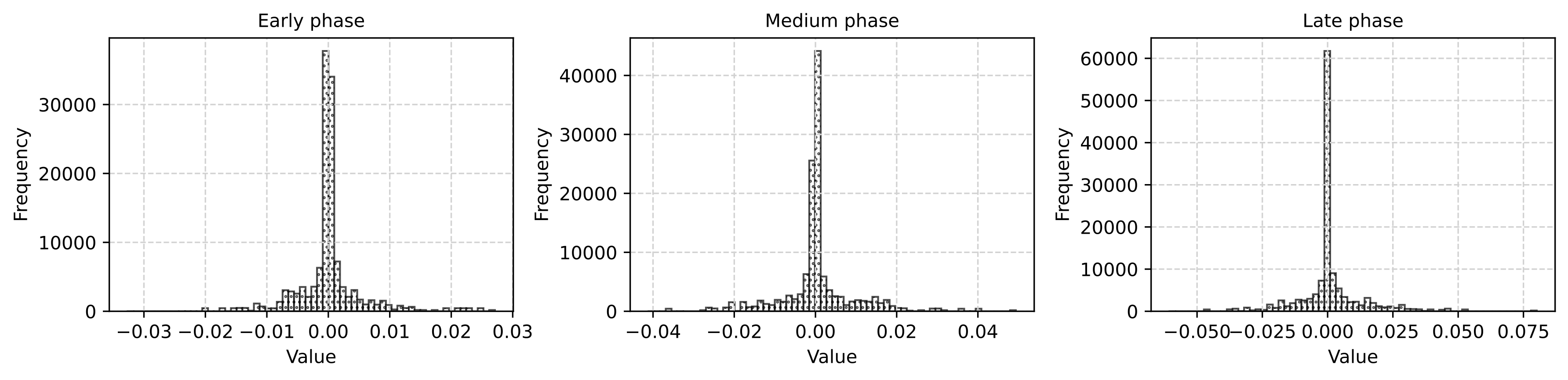}
\caption{Data distribution of representative EMB tables in different phases on Criteo Terabyte dataset.}
\vspace{-5mm}
\label{fig:distribution_phases}
\end{figure*}

\textbf{Compression Performance across Training Phases.} Based on Figures \ref{kaggle_global_cr} and \ref{kaggle_cons_cr}, we observe that our compressor performs effectively throughout all training phases, maintaining a consistently high compression ratio. Implementing an error-bound decay to enhance the quality of compressed data results in only a slight decrease in the compression ratio. This stable compression ratio can be attributed to two main factors. Firstly, the Vector-based LZ encoder, which relies on pattern matching, maintains its effectiveness due to the presence of identifiable patterns in each batch, a factor that remains constant regardless of the training phase. Secondly, the uniformity in data distribution across the training, as depicted in Figure \ref{fig:distribution_phases}, ensures consistent compression performance.

\subsection{Evaluation of Compression Optimization}
Finally, we evaluate the performance of our optimized compression compared to the non-optimized solution. 

\textbf{LZ Fine-tuning.} First, we evaluate our optimized compressor with different LZ fine-tuning window sizes. Table \ref{table:LZ fine-tuning} shows the compression ratio and throughput on DLRM data with varying window sizes. Generally, larger window sizes result in more pattern matches. For the Criteo Terabyte dataset, we observe that the overall compression ratio with the window sizes of 128 and 255 are 3.9$\times$ and 5.2$\times$ higher, respectively, compared to the baseline window size of 32. On the Criteo Kaggle dataset, the difference between the window sizes of 128 (1.52$\times$) and 255 (1.54$\times$) is negligible. In instances of small batch sizes, our vector-based LZ is less beneficial with the increase in window size. This is due to the proportion between the sliding window size and the volume of data. In the model of the Criteo Kaggle dataset, for example, $Batch Size$ is 128 by default and can be fully covered by one sliding window. The scaled compression ratios of the Criteo Terabyte dataset demonstrate that increasing the window size does not linearly increase the compression ratio. This phenomenon is attributed to the unbalanced frequency of queries. EMB vectors with very high frequency can be matched with an appropriate window size, whereas EMB vectors with low frequency require a window size that exceeds hardware limitations and is inefficient.

\begin{table}[t]
\centering\sffamily\footnotesize
\caption{Compression ratio improvement of fine-tuned LZ encoder with different window sizes.}
\label{table:LZ fine-tuning}
\begin{tabular}{ @{} l c c c c @{} }
\toprule
Window Size & 32& 64 & 128 & 255 \\ 
\midrule
Criteo Kaggle & 1$\times$ & 2.21$\times$ & 3.89$\times$ & 5.23$\times$ \\ 
Criteo Terabytes & 1$\times$ & 1.47$\times$ & 1.52$\times$ & 1.54$\times$ \\ 
\bottomrule
\end{tabular}
\end{table}

\textbf{Buffer Optimization.} Second, we evaluate our buffer optimization in both compression and decompression processes. We split the EMB vectors into chunks, with the number of chunks equal to the RANK in distributed training, which reflects the scalability of the training process. Figure \ref{buffer_optimizationp} illustrates the compression speedup across different EMB vector sizes, with the number of chunks ranging from 2 to 16. The term 'single\_comp' denotes our solution. The results indicate that our design achieves higher speedups with an increased number of chunks. According to our evaluation, our optimizations achieve a maximum speedup of 2.04$\times$.

Our proposed buffer optimization technique performed as expected. When evaluating speed-up across different data chunk sizes, we observed that with 8MB data blocks, the performance is 1.86$\times$ better than with 64MB blocks. As discussed previously, for limited data sizes, the volume of an individual chunk is too small to achieve optimal GPU utilization. The bottleneck in compression for these small chunks arises from frequent kernel launches rather than from the compression process or memory copying itself. Conversely, for larger data volumes, the advantage of buffer optimization becomes less significant because each chunk's larger volume allows the GPU to attain higher utilization, even when compressing chunks sequentially.

\begin{figure}[t]
\centering
\includegraphics[width=\linewidth]{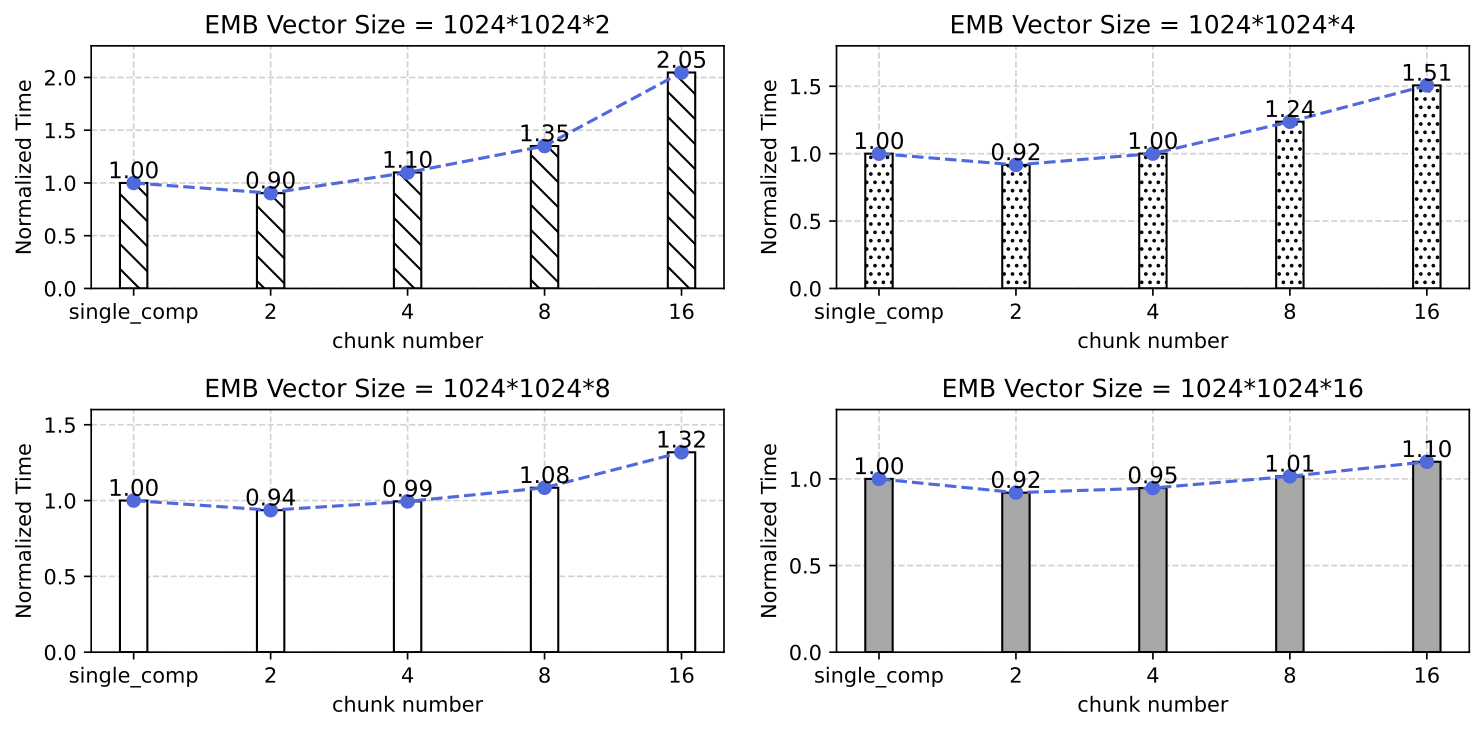}
\caption{Normalized time of our work with and without buffer optimization for different EMB vector sizes. `single\_comp` denotes our solution, while the chunk number indicates how many chunks the original EMB vector is equally partitioned and compressed.}
\label{buffer_optimizationp}
\end{figure}

\section{Related Work}
\label{sec:related}

A variety of research efforts have sought to expedite DLRM training, generally falling into three categories: embedding compression, embedding caching, and low-precision training. 

\textbf{Embedding Compression.} Yin \textit{et al.}'s TT-Rec \cite{yin2021ttrec} and Wang \textit{et al.}'s EL-Rec \cite{el-rec} utilize Tensor Train decomposition to reduce memory consumption in Embedding Tables, aiming to lessen resource usage in constrained environments. Despite their advantages, these model-compression techniques face challenges: firstly, they may not always ensure convergence or maintain high model accuracy; secondly, they can introduce significant computational overheads. Specifically, the recovery of embedding vectors for each batch necessitates extra matrix multiplications. Given the often long and narrow shape of these matrices, such operations can prove highly inefficient on GPUs. Furthermore, it is important to highlight that these techniques are complementary to our approach. Essentially, our method could be combined with model compression in DLRM training to further enhance performance in scenarios with limited resources.

\textbf{Embedding Caching.} Strategies like Pattern-Aware Sparse Communication by He \textit{et al.} \cite{24he_pattern_aware}, cDLRM by Balasubramanian \textit{et al.} \cite{10.1145/3460231.3474246}, and a heterogeneous SmartNIC system by Guo \textit{et al.} \cite{guo2023software} aim to alleviate caching overhead. Unlike these approaches that may require additional memory for data copies or hardware support for caching, our compression technique avoids extra storage demands for heterogeneous embedding table access. Moreover, maintaining cache involves computational costs, such as updating cache entries and ensuring cache coherence, which our method does not incur.

\textbf{Low-precision Training.} Exploring low-precision data types, as proposed by Rouhani \textit{et al.} with the new datatype MX\cite{0.02threshold} and mixed-precision strategies by Yang \textit{et al.}\cite{yang2020mixedprecision}, is another direction of research. These methods, though direct, offer limited compression benefits due to their fixed compression ratio and lack of fine-tuning capabilities, providing a coarse granularity control over data compression. Our strategy, in contrast, allows for a smooth adjustment of error bounds, offering a more flexible and efficient solution to communication reduction in DLRM training.

\textbf{General Compression-accelerated Communication.}
In addition, several studies have explored leveraging compression to boost communication speeds more generally. Zhou \textit{et al.} have contributed a series of works \cite{DBLP:conf/ipps/ZhouAXSASP23}, \cite{DBLP:conf/hipc/ZhouASSP22} that emphasize enhancing collective communication efficiency through compression. Similarly, Ramesh \textit{et al.} introduced Efficient Pipelined Communication Schemes \cite{DBLP:conf/hipc/RameshZSASP22}, aimed at minimizing blocking time and maximizing bandwidth utilization.

Our approach, however, stands out from these efforts in a couple of key ways. Firstly, unlike the aforementioned studies, our method is tailor-made for DLRM applications, incorporating adaptive error-bound adjustments that are absent in general compression techniques. These general approaches often provide a low-level interface that may prove challenging to finely tune in real-world scenarios. Secondly, our compression algorithm is specifically optimized for DLRM training, offering both higher compression throughput and ratios compared to existing GPU compressors—let alone CPU compressors, which generally deliver significantly lower throughput. Overall, our method provides an integrated end-to-end solution specifically for enhancing DLRM training communication, unlike the general methods that only offer a compression-accelerated communication library or tool without tailoring to the specific needs of the application.

\section{Conclusion and Future Work}
\label{sec:conclusion}
In this paper, we introduce a method that employs error-bounded lossy compression to reduce the communication data size and accelerate DLRM training. We develop a novel error-bounded lossy compression algorithm to achieve hybrid by hybridizing our optimized LZ encoder and entropy encoder on GPU. Moreover, we introduce a dual-level adaptive strategy for error-bound adjustment, spanning both table-wise and iteration-wise aspects, to balance the compression benefits with the potential impacts on accuracy. We further optimize our compressor for PyTorch tensors on GPUs, minimizing compression overhead. The evaluation shows that our method achieves an 8.6$\times$ all-to-all communication speedup and 1.38$\times$ end-to-end training speedup with a minimal accuracy impact.

In the future, we plan to further refine our system to reduce compression overhead by employing strategies like kernel fusion on GPUs and seamlessly integrating (de)compression processes with communication libraries such as NCCL. Additionally, we aim to develop a more advanced and automated approach for offline selection of a fixed global error-bound and for online error-bound adjustments.

\section*{Acknowledgment}

The work is supported by the Meta Research Award for ``AI System Hardware/Software Codesign.'' 
Hao Feng, Boyuan Zhang, Fanjiang Ye, and Dingwen Tao's work on this project was supported by the National Science Foundation (Grant Nos. 2312673, 2247080, 2303064, 2326494, and 2326495). Dingwen Tao was also supported by the National Natural Science Foundation of China (Grant Nos. 62032023 and T2125013) and the Innovation
Funding of ICT, CAS (Grant No. E461050).

\renewcommand*{\bibfont}{\normalfont\small}
\printbibliography[]

\end{document}